\documentclass[conference,compsoc]{IEEEtran}
\usepackage{cite}
\usepackage{comment}
\usepackage{amsmath,amssymb,amsfonts}
\usepackage{algorithmic}
\usepackage[font=normalsize]{subfig} 
\usepackage[style=base]{caption}
\usepackage{graphicx}
\usepackage[normalem]{ulem}
\usepackage{textcomp}
\usepackage{xcolor}
\usepackage{mathtools}
\usepackage{url}
\usepackage{float}
\newcommand{\eatme}[1]{ }
\def\BibTeX{{\rm B\kern-.05em{\sc i\kern-.025em b}\kern-.08emhttps://www.overleaf.com/project/6337481974f7760998146f56
    T\kern-.1667em\lower.7ex\hbox{E}\kern-.125emX}}

\usepackage{xparse}
\NewDocumentCommand\N{sm}{\mathcal{N}\IfBooleanT#1{^{\ast}}_{#2}}

\makeatletter
\NewDocumentCommand{\@Coefficients}{m}{\text{\ttfamily\upshape #1}}
\newcommand\uMultilevelCoefficients{\@Coefficients{u\char`_mc}}
\newcommand\newMultilevelCoefficients{\@Coefficients{\~{f}\char`_mc}}
\makeatother

\begin{document}
\IEEEoverridecommandlockouts

\title{Scalable Hybrid Learning Techniques for Scientific Data Compression
\vspace{-0.5cm}}

\author{
    \IEEEauthorblockN{Tania Banerjee\IEEEauthorrefmark{1}, Jong Choi\IEEEauthorrefmark{2}, Jaemoon Lee\IEEEauthorrefmark{1}, Qian Gong\IEEEauthorrefmark{2}, Jieyang Chen\IEEEauthorrefmark{2},\\
    Scott Klasky\IEEEauthorrefmark{2}, Anand Rangarajan\IEEEauthorrefmark{1}, 
    Sanjay Ranka\IEEEauthorrefmark{1}}
    \IEEEauthorblockA{\IEEEauthorrefmark{1}University of Florida, USA
    }
    \IEEEauthorblockA{\IEEEauthorrefmark{2}Oak Ridge National Laboratory, USA
    }
    \vspace{-0.5cm}
    \thanks{Contact e-mail: sranka@ufl.edu. This research was funded by the DOE (Grant No. DE-SC0022265).}
}

\maketitle

\begin{abstract}
Data compression is becoming critical for storing scientific data because many scientific applications need to store large amounts of data and post process this data for scientific discovery. Unlike image and video compression algorithms that limit errors to primary data, scientists require compression techniques that accurately preserve derived quantities of interest (QoIs). This paper presents a physics-informed compression technique implemented as an end-to-end, scalable, GPU-based pipeline for data compression that addresses this requirement.
Our hybrid compression technique combines machine learning techniques and standard compression methods. Specifically, we combine an autoencoder, an error-bounded lossy compressor to provide guarantees on raw data error, and a constraint satisfaction post-processing step to preserve the QoIs within a minimal error (generally less than floating point error).

The effectiveness of the data compression pipeline is demonstrated by compressing nuclear fusion simulation data generated by a large-scale fusion code, XGC, which produces hundreds of terabytes of data in a single day. Our approach works within the ADIOS framework and results in compression by a factor of more than 150 while requiring only a few percent of the computational resources necessary for generating the data, making the overall approach highly effective for practical scenarios.

\end{abstract}

\begin{IEEEkeywords}
ITER, XGC data compression, autoencoder, machine learning, MGARD, SZ, ZFP, Summit
\end{IEEEkeywords}

\section{Introduction}

The volumes and velocities of data generated by scientific applications have continued to outpace the growth of computing power and network and storage bandwidths and capacities~\cite{foster2021online}. This growth is also seen in next-generation experimental and observational facilities, making data reduction (or compression) an essential functionality. Data reduction necessitates the development of trustworthy algorithms so that scientists can rely on the reduced data, and more so because, in many cases, it will be impractical to retain the original, nonreduced data. Thus, we need to quantify the error of the reduced data. Further, data reduction approaches should seamlessly integrate into data management workflows and storage at leading computing facilities. 

Lossy data reduction or compression techniques have been successfully applied to image and video processing applications \cite{grois2013performance}. Still, these techniques are not directly applicable to scientific applications because these applications also need to quantify the
error for Quantities of Interest (QoIs), which are derived quantities from the primary data.

Limiting these errors with realistic numerical bounds is essential for scientists' confidence in lossy data reduction. This paper presents an end-to-end algorithmic and software pipeline for data compression that guarantees error bounds on primary data and quantities of interest for the XGC application~\cite{chang2009whole}---a large-scale nuclear fusion simulation code---which produces hundreds of terabytes in a single day. The QoIs of importance for this application include moments of the particle histogram with the number of QoIs being $\mathcal{O}(1)$ in the cardinality of the primary data---the particle histograms in the velocity space. 
\begin{figure*}[htp]
    \centering
    \includegraphics[width=0.565\textwidth, trim={0cm 2cm 4.1cm 1cm},clip]{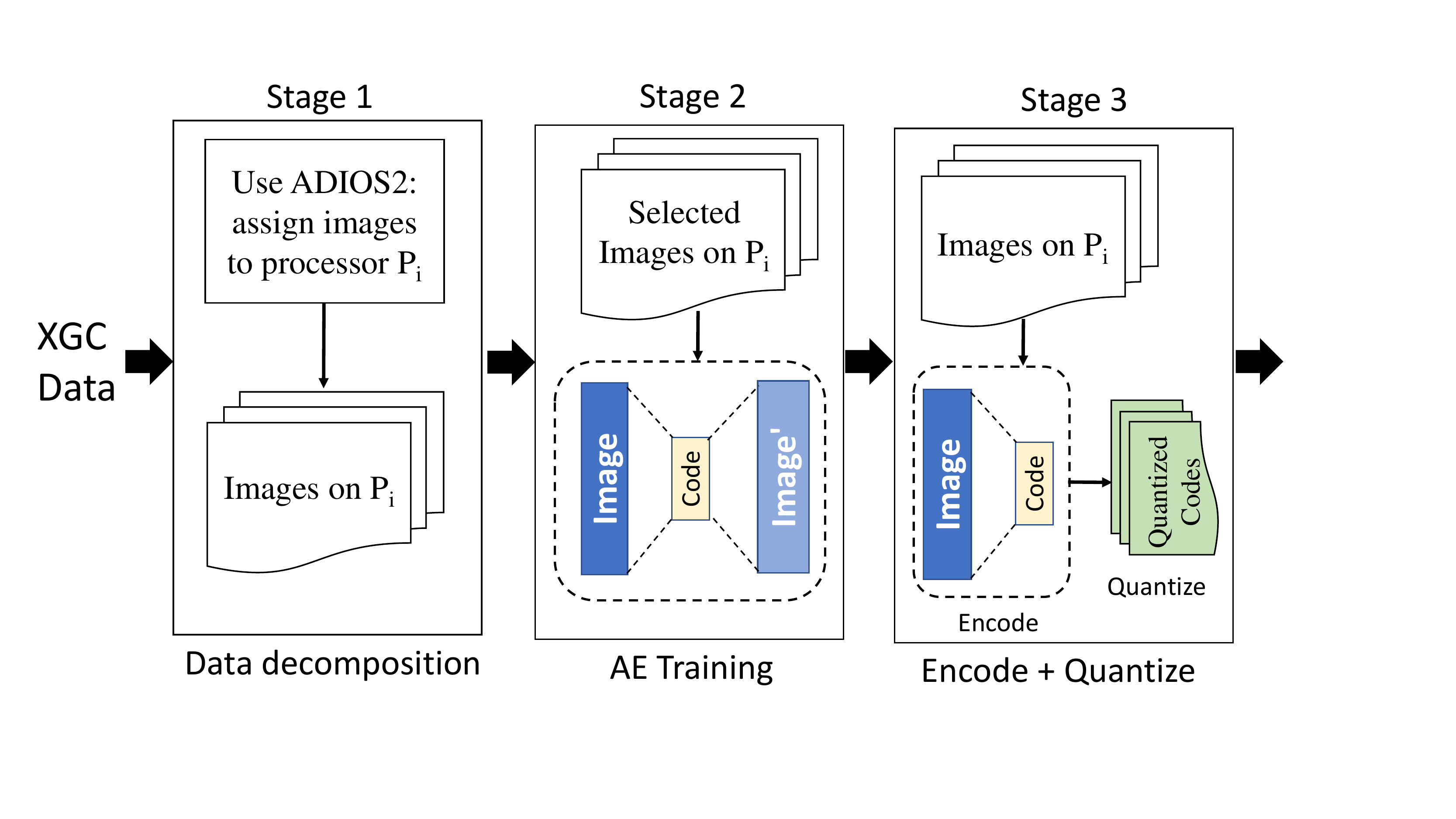}
    \includegraphics[width=0.42\textwidth, trim={6.85cm 1.5cm 5cm 2cm},clip]{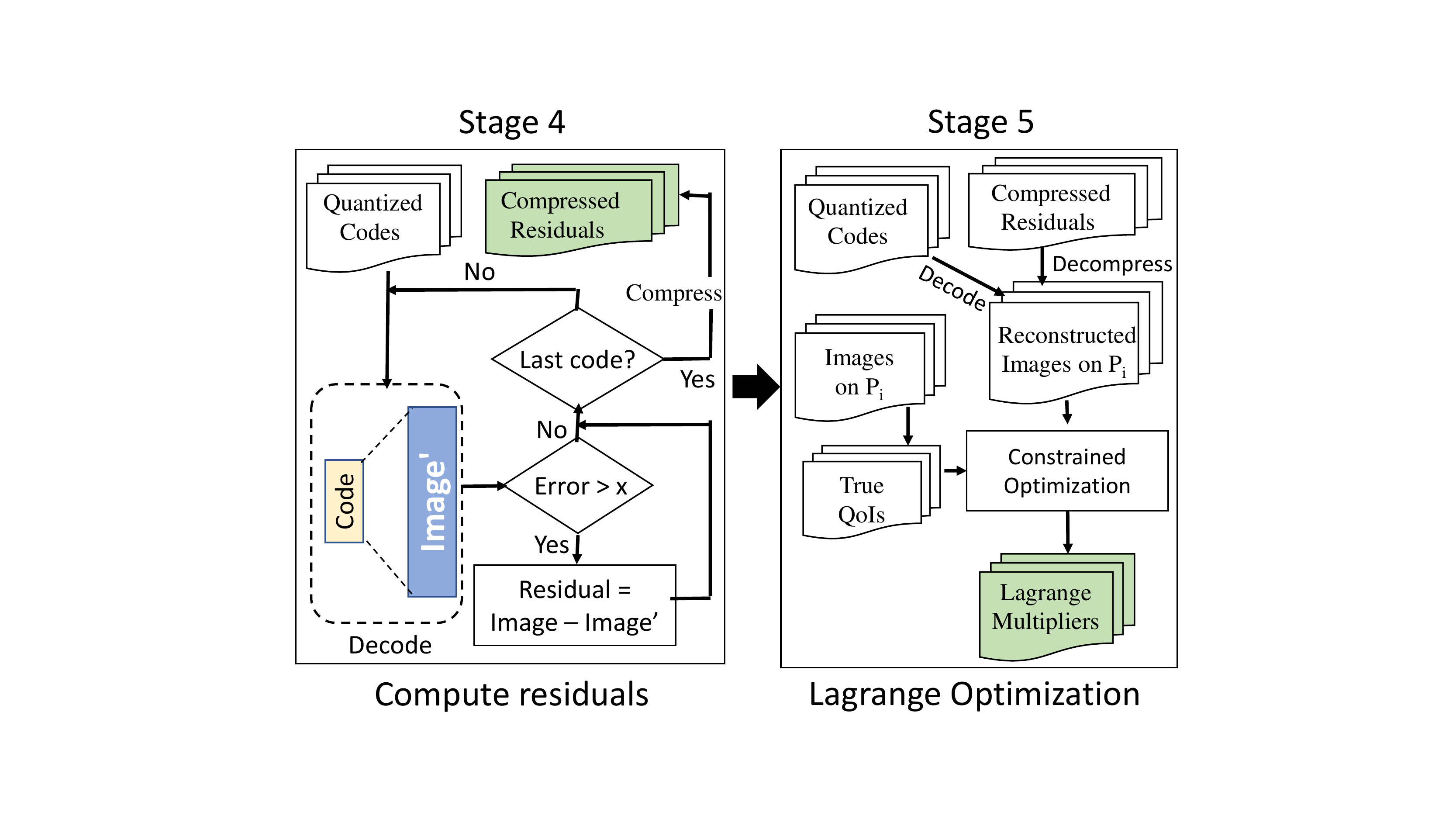}
    \caption{Data compression pipeline in processor $P_i$. The input to the pipeline is XGC data, while the outputs are quantized codes, compressed residuals, and Lagrange multipliers (in green boxes), along with the autoencoder model weights. An image is a particle histogram in 2D velocity space. XGC data contains millions of such images and is generated from simulation.}
    \label{fig:pipeline}
\end{figure*}

Figure~\ref{fig:pipeline} shows the key stages of our pipeline described as follows:
\begin{itemize}
    \item 
Stage 1 decomposes the data using domain decomposition into small subdomains and assigns them to processors. Each processor processes data in a subdomain independently to achieve a high level of parallelism. 
\item 
Stage 2 trains an autoencoder (AE) for data compression.
\item
Stage 3 uses the AE to encode data and quantizes the encoded data to further reduce its size.
\item
Stage 4 applies residual processing using a standard lossy compressor which guarantees error bound on the primary data. Stage 4 processes the residuals selectively only on the part of data where the PD error is above a user-provided threshold. Thus, error guarantees in the form of bounded reconstruction errors are achieved overall.
\item
Stage 5 uses a post-processing optimization technique---essentially a nonlinear projection of the PD into a subspace spanned by the $\mathcal{O}(1)$ QoI constraints expressed by Lagrange multipliers, $\lambda$s. Unlike directly storing the QoIs, this nonlinear projection operation brings the compressed PD in line with the constraints. The $\lambda$s generated can be further quantized or truncated in this step to increase the compression level. 
\end{itemize}
We demonstrate the effectiveness of the pipeline by compressing simulation data generated on the ITER simulation for XGC~\cite{iter}. XGC uses ADIOS-2~\cite{suchyta2022exascale} to achieve high-performance, self-describing parallel I/O. ADIOS can reach over 1 TB/s on the Summit supercomputer GPFS file system.

 We show that our approach can compress the data by two orders of magnitude with guarantees on the PD and QoI errors. In particular, the post-processing optimization technique achieves high accuracy on the QoIs. For example, the QoIs are accurate by up to $10^{-14}$ when the $\lambda$s are stored as double precision floating point numbers and up to $10^{-8}$ when the $\lambda$s are stored as single precision floating point numbers. The PD error is a user-tunable  parameter based on desired accuracy bounds. 
 
 The compression can be conducted by setting aside computational resources. It should not impact the simulation performance because the amount of resources required for the compression is a few percent of the necessary resources for the XGC simulation. These characteristics of our approach make it highly scalable and practical for applying on-the-fly compression while guaranteeing errors on QoIs that are critical to scientists analyzing the datasets or for downstream applications that utilize the data. To the best of our knowledge, this is the first work demonstrating an end-to-end, parallel, and scalable data reduction pipeline capable of compressing petabytes of data while guaranteeing error bounds on PD and preserving the known QoIs.
 
A preliminary version of our work appeared in~\cite{hipc2022}. The key contributions in this extended version over the work presented in \cite{hipc2022}  are as follows:
 \begin{enumerate}
     \item Addition of the neural network based autoencoder and quantization as described in Stages 2, 3, and 4. The approach in \cite{hipc2022} was limited to only using the state-of-the-art lossy compressor, MGARD, for primary data compression. The autoencoder techniques were shown to be effective in our earlier paper for small datasets \cite{jaemoon2} executing sequentially.
     \item Novel domain decomposition and incremental training approaches that keep the overall approach highly parallel.
     \item
     Detailed experimental results on both ion and electron data. The prior work was limited to ions.
 \end{enumerate}
This paper provides extensive experimental results on all the key parameters necessary to make our approach useful and practical for large data sets.
 
 From an implementation point of view, our implementation is based on PyTorch C++, which improves the code timing performance by effectively using the GPUs, making the code production ready. Further, this paper develops a scalable MPI-based compression pipeline with efficient I/O. This implementation infrastructure makes it viable to integrate the pipeline in this paper with scientific applications. To the best of our knowledge, this is the first software pipeline to achieve parallelized compression on PD (while satisfying QoIs), making it viable for integration with scientific applications.
 
 The rest of the paper is organized as follows. Section~\ref{sec:app} presents the XGC application, Section~\ref{sec:method} explains our data compression pipeline and the associated components, Section~\ref{sec:expt} presents our experiments, Section~\ref{sec:related} presents related work, and finally, we conclude in Section~\ref{sec:concl}.


\section{XGC Application\label{sec:app}}
The X-point Gyrokinetic Code (XGC)~\cite{chang2008spontaneous,ku2009full} describes the kinetic transport properties of tokamaks. According to the U.S. Department of Energy (DOE), a tokamak is a machine that confines a plasma using magnetic fields in a doughnut-shaped domain, also known as a torus. A plasma is a superheated gas that is a soup of positively charged particles (ions) and negatively charged particles (electrons). The tokamak is believed by fusion scientists to be the most promising approach to plasma confinement for fusion power plants. However, the plasma is difficult to sustain because the presence of strong particle and energy sources and sinks near the edge of a tokamak plasma disturbs the plasma's thermal equilibrium, invalidating the underlying equations in fluid mechanics. XGC was developed to describe the kinetic transport phenomenon in this complicated region. 

XGC is a massively parallel, Particle-In-Cell (PIC) code that simulates trillions of particles (electrons, ions, neutrons, etc.) in parallel to resolve complex plasma properties in a tokamak. The particles in XGC evolve in a time-marching fashion according to governing, five-dimensional, gyrokinetic equations of motion and then interpolated or ``gathered'' to discrete mesh points for solving electric and magnetic field equations which factor into the particle motion in subsequent steps. Particles are regularly histogrammed to form particle distribution functions, $F$, in five-dimensional space: three in physical space and two in velocity space. Effectively $F$ can be represented as two dimensional tensors (or images) in a three dimensional grid. The size of $F$ data ranges from a few GBs up to tens of TBs per iteration, depending on the mesh resolution. 

Due to I/O bottlenecks on the supercomputer and storage limitations, various reductions are often applied to $F$ to form physically relevant quantities to output, such as density ($n$) and temperature ($T$), instead of writing out the entire $F$ data. But because these always involve a loss of information, methods to quickly and efficiently reduce or compress $F$ data (and in the future, the particle data itself) are desirable to enable scientists to capture richer physics from these simulations. 

\begin{figure}[ht!]
    \centering
    \includegraphics[width=0.7\columnwidth]{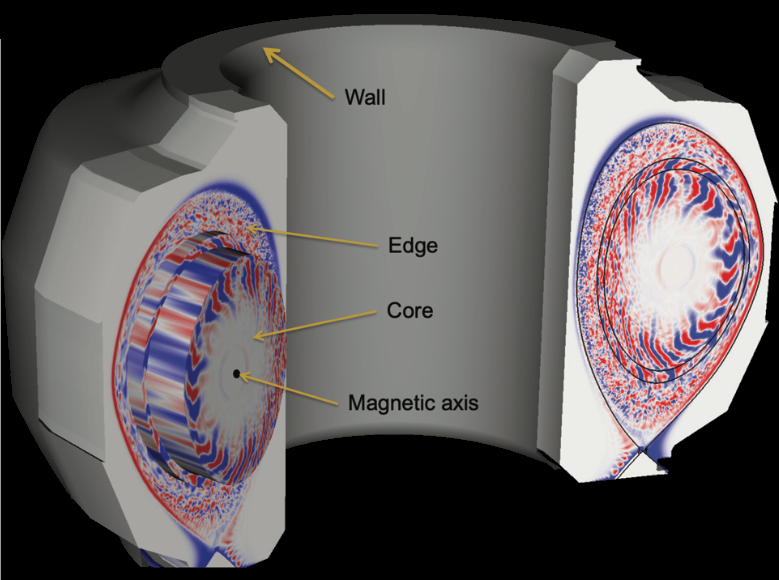}\\
    \includegraphics[trim={30 20 63 20},clip,width=0.265\columnwidth]{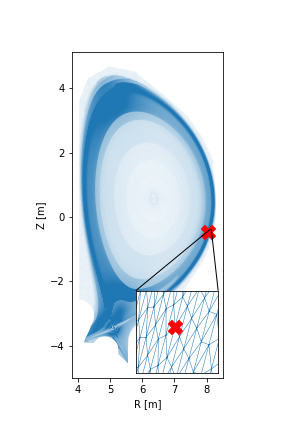}
    \includegraphics[width=0.72\columnwidth]{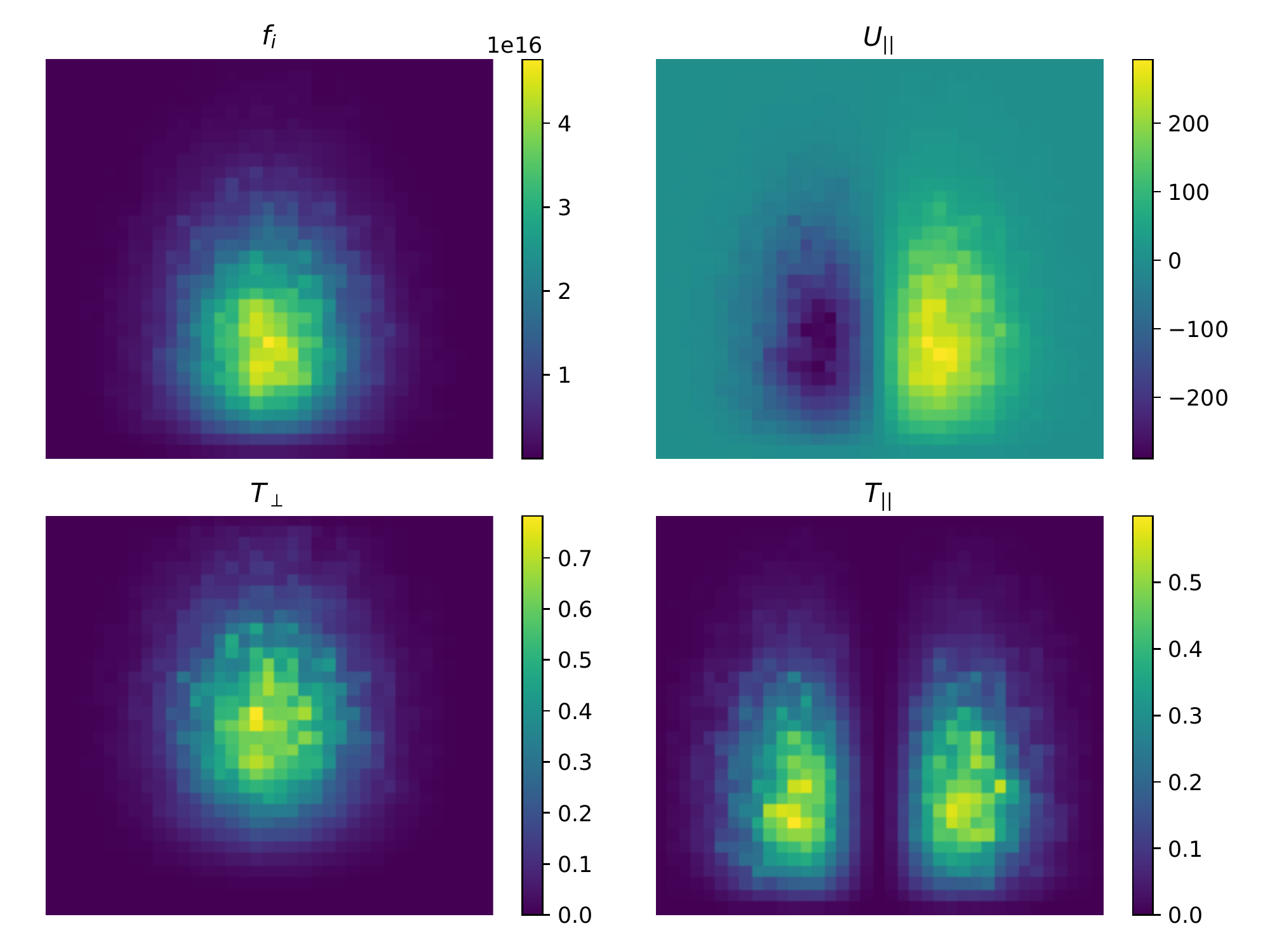}
    \caption{
    A cross-sectional structure of tokamak~\cite{dominski2021spatial} that XGC simulates (top) and an example of XGC ITER mesh and the $F$ data and their derived quantities  (bottom). The marker shows the location of $F$ data collected at step 200.
    }
    \label{fig:xgc2}
\end{figure}

This paper uses data generated from ITER simulations~\cite{d2017fusion, iter} to demonstrate the compression pipeline. ITER is one of the world's largest tokamak, being designed by thousands of scientists worldwide. ITER is being developed to produce 500 MW of power and has ten times the plasma volume as the largest operating fusion machine. ITER simulations produce over $160$~GB of $F$ data per 30 seconds. Hence, data compression is an absolute necessity to store these data while at the same time ensuring that the various physical properties derived from the data remain accurate in the reconstructed data.

There are four QoIs that need to be preserved and two other QoIs that are derived from the first four. The four QoIs are density, parallel flow velocity, perpendicular temperature, and parallel temperature, represented as $n$, $u_{\parallel}$, $T_{\perp}$ and $T_{\parallel}$, respectively. The two derived QoIs are flux surface averaged density and flux surface averaged temperature, represented as $n_{0}^{avg}$ and $T_{0}^{avg}$, which are, respectively, the averages of density and temperature QoIs~\cite{osti_1564081}.

The physics equations for computing the QoIs are presented here for the sake of completeness. These equations were used for derivations of the $\lambda$s. 
\begin{equation}\label{eq:den}
n\left(\boldsymbol{x},t\right)=\int f\left(\boldsymbol{x},\boldsymbol{v},t\right)\textrm{vol}\left(x,\boldsymbol{v}\right)dv_{\perp}dv_{\parallel},
\end{equation}
\begin{equation}\label{eq:upara}
u_{\parallel}\left(\boldsymbol{x},t\right)=\frac{\int f\left(\boldsymbol{x},\boldsymbol{v},t\right)\textrm{vol}\left(\boldsymbol{x},\boldsymbol{v}\right)v_{\parallel}\left(\boldsymbol{v}\right)dv_{\perp}dv_{\parallel}}{n\left(\boldsymbol{x},t\right)},
\end{equation}
where $t$ represents the timestep, $\textrm{vol}\left(\boldsymbol{x},\boldsymbol{v}\right)$ is the volume of the $x^{th}$ mesh node, and $v_{\parallel}$ is the parallel velocity. $T_{\perp}$ and $T_{\parallel}$ are the integrals of perpendicular and parallel kinetic energy, $mv^{2}$,
\begin{equation}\label{eq:tperp}
T_{\perp}\left(\boldsymbol{x},t\right)=\frac{1}{2}\frac{\int f\left(\boldsymbol{x},\boldsymbol{v},t\right)\textrm{vol}\left(x,\boldsymbol{v}\right)mv_{\perp}^{2}\left(\boldsymbol{v}\right) dv_{\perp}dv_{\parallel}}{n\left(\boldsymbol{x},t\right)},
\end{equation}
\begin{equation}\label{eq:tpara}
\begin{aligned}
& T_{\parallel}\left(\boldsymbol{x},t\right)= \\
& \frac{1}{2}\frac{\int f\left(\boldsymbol{x},\boldsymbol{v},t\right)\textrm{vol}\left(\boldsymbol{x},\boldsymbol{v}\right)m\left(v_{\parallel}\left(\boldsymbol{v}\right)-u_{\parallel}\left(\boldsymbol{x},t\right)\right)^{2} dv_{\perp}dv_{\parallel}}{n\left(\boldsymbol{x},t\right)},
\end{aligned}
\end{equation}
where, $f$ denotes any reconstruction of $F$ data which strictly satisfies the QoI constraints, $m$ is the mass of the charged particles, and $v_{\perp}$ is the velocity perpendicular to the direction of the magnetic field. 
\eatme{
XGC uses ADIOS for data management to input/output large-scale physics data, including $F$ data, focused on running on an HPC system.
}

\section{Data Compression Pipeline\label{sec:method}}
A tokamak simulation generates massive amounts of data that we intend to compress. A tokamak has a toroidal shape, and simulations happen across multiple cross-sections of the torus, referred to as planes. Each plane is discretized using a triangular mesh overlay. The particles present at each mesh node are histogrammed based on their velocity with discretized velocity components along directions parallel and perpendicular to the magnetic field. The input data to our compression pipeline are the particle histograms in the two-dimensional velocity space at the mesh nodes on each plane. Therefore, the size of the data depends upon the resolution of the underlying mesh. An observation that has led to considerable improvement in our AE training time, as we will see later, is that the particle distribution at mesh nodes is similar across planes. The data compression pipeline compresses the generated data.
\begin{figure}[tbp]
    \centering
    \subfloat[Row-wise data decomposition]
    {\includegraphics[width=0.55\textwidth, trim={3cm 5cm 0cm 1cm},clip]{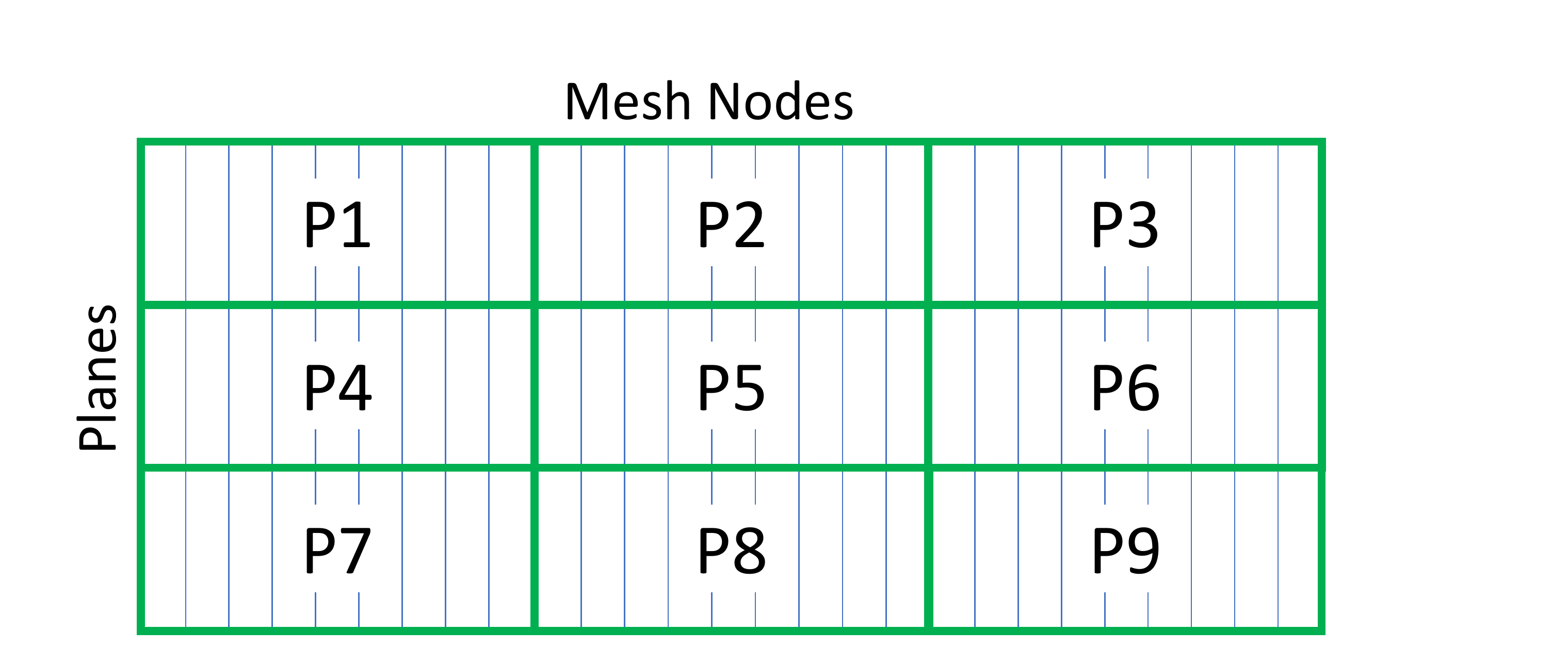}}
    \quad
    \subfloat[Column-wise data decomposition]
    {\includegraphics[width=0.52\textwidth, trim={2cm 4cm 3cm 2cm},clip]{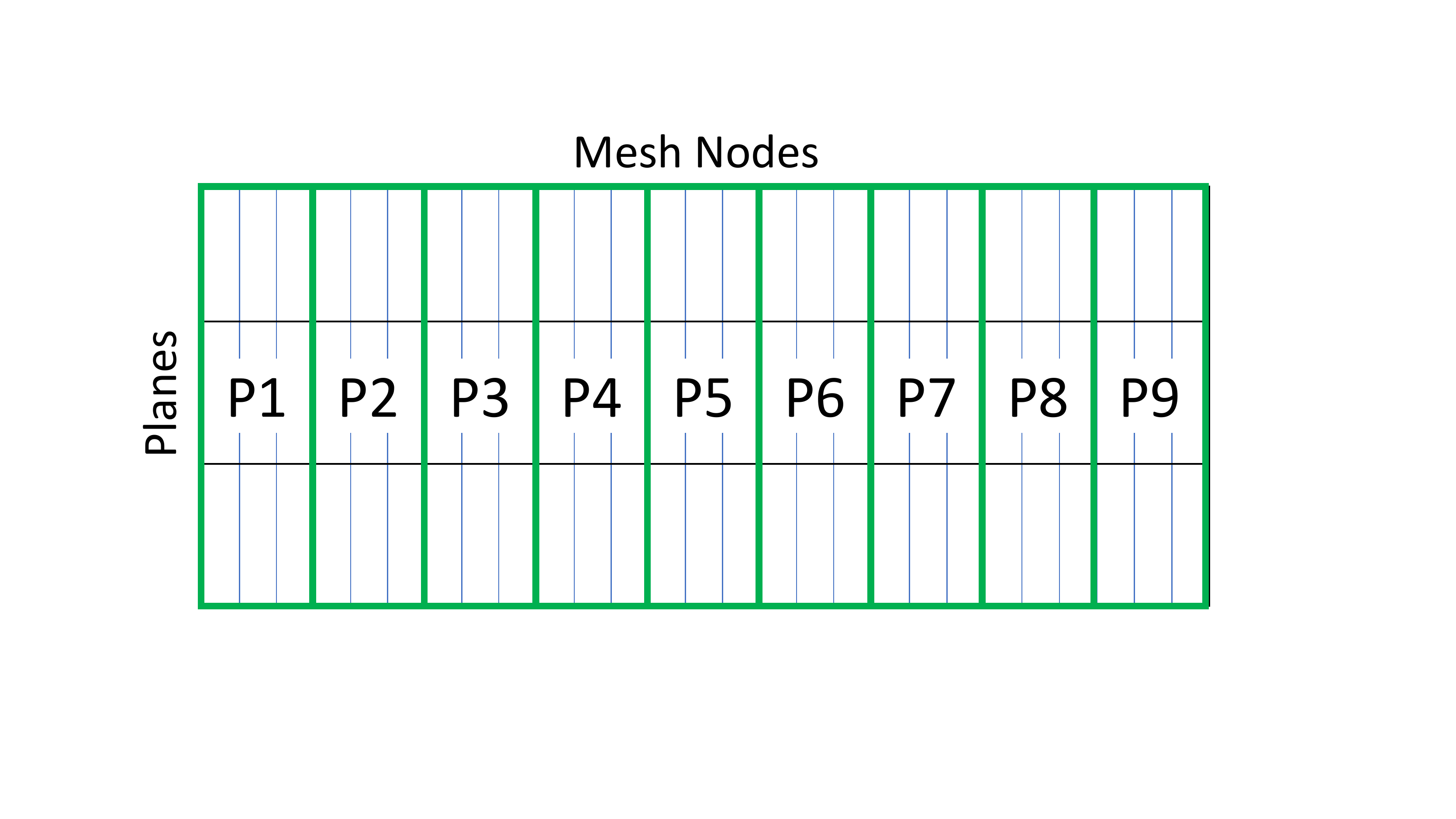}}
    \caption{\small Data decomposition scheme for a XGC dataset with three planes. In this example, there are nine processors, $P1, P2, \cdots, P9$ available for processing the mesh nodes. Each mesh node consists of a two-dimensional particle histogram in the ($V_x$,~$V_y$) velocity space. (a) Row-wise decomposition assigns mesh nodes in a plane to a subset of three processors, and each processor gets data from one plane only;  (b) Column-wise decomposition assigns mesh nodes in a plane to all processors and each processor gets data from all planes.}
    \label{fig:datadecomp}
\end{figure} 
The compression pipeline has five main stages.
The first stage decomposes the data using domain decomposition into small subdomains and assigns them to processors.
The second stage trains a simple autoencoder to map the data to a bottleneck layer. The third stage uses the trained autoencoder to encode the input data and also apply a quantization scheme to the encoded data for further reduction. The fourth stage employs a residual processing technique to guarantee error bounds on the primary data. The fifth and final stage uses a post-processing optimization technique---essentially a nonlinear projection of the primary data into a subspace spanned by the $\mathcal{O}(1)$ QoI constraints expressed by the Lagrange multipliers, $\lambda$s. These five stages are now described in detail.

{\it Stage 1: Domain decomposition:} Domain decomposition is a technique to divide the input tensor into smaller chunks and distribute these to the parallel processors for compressing each piece in parallel. 
\eatme{ The data are read and decomposed efficiently using the ADIOS framework.} This paper uses two domain decomposition techniques: a row-wise and a column-wise decomposition of the input tensor. Figure~\ref{fig:datadecomp} illustrates the difference between row-wise and column-wise data decomposition schemes. In a row-wise data decomposition scheme, mesh nodes on each XGC plane (or row) are processed using a subset of processors. For example, processors P1, P2, and P3 in Figure~\ref{fig:datadecomp}(a) processes data in the first plane. The mesh nodes on each plane are divided approximately equally among the processors responsible for processing that plane. In a column-wise data decomposition scheme, mesh nodes on each XGC plane are processed using all processors. As a result, a processor is assigned a subset of data from all the planes. For example, processor P1 in Figure~\ref{fig:datadecomp}(b) processes data from planes 1, 2, and 3. Data decomposition makes our approach highly scalable because the compression steps can be applied independently to each subdomain.

\begin{figure}[htp]
    \centering
    \includegraphics[width=1\columnwidth]{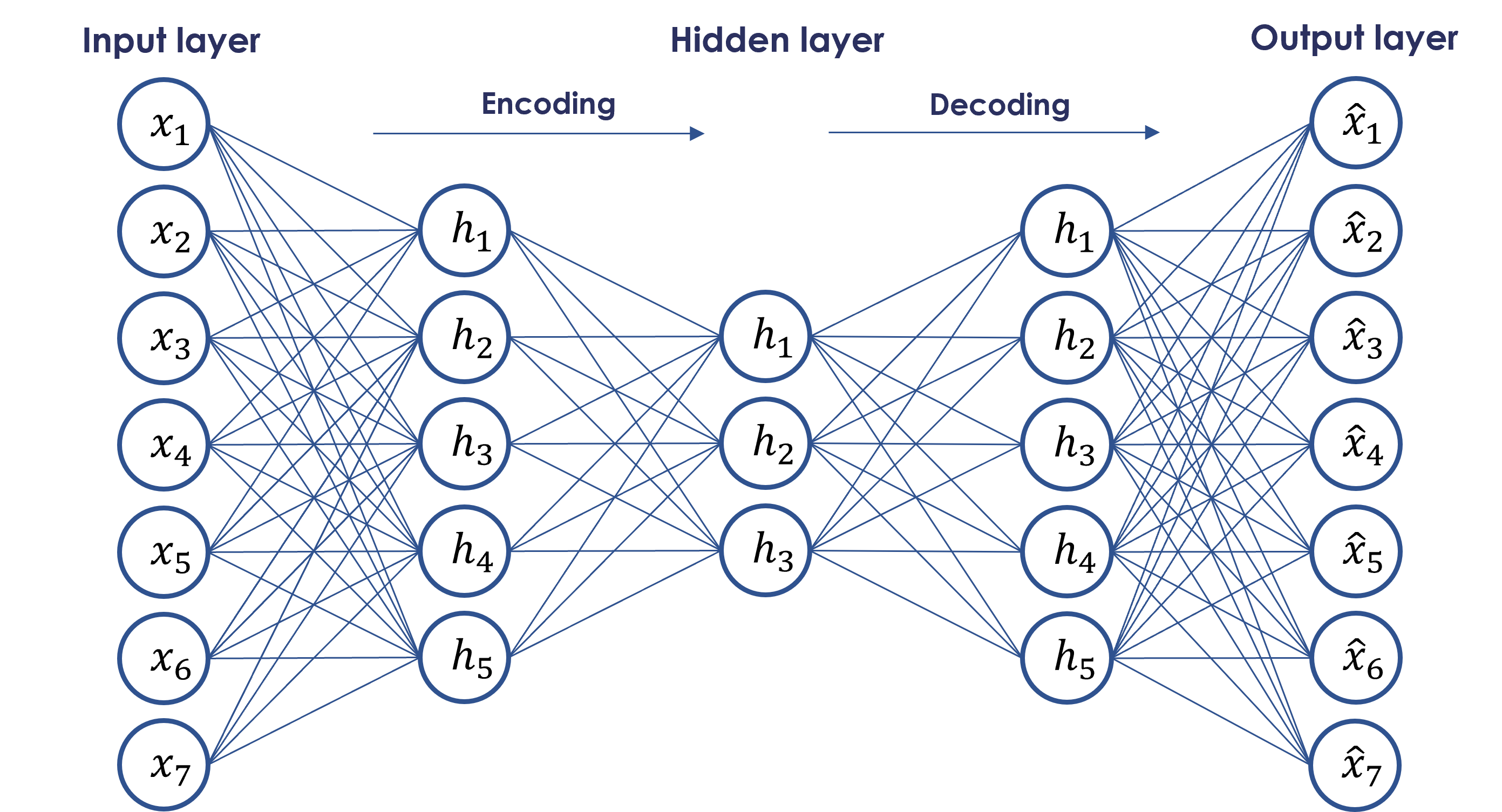}
    \caption{An autoencoder with input, output, and the hidden layers, which include a bottleneck layer.}
    \label{fig:autoencoder}
\end{figure}

{\it Stage 2: Training an Autoencoder:}
An autoencoder (AE) is a neural network that is trained to replicate the input at its output while going through a bottleneck layer. Broadly, an AE has three components: encoder, bottleneck, and decoder.  An AE may have several hidden layers, and the bottleneck is the final hidden layer that learns a code to represent the input. The encoded coefficients (also known as the latent space) in the bottleneck are obtained using standard loss measures in the AE. As a result, the PD errors are minimized by the AE while learning a good codec. The loss function of the AE network is a function of the PD and the reconstructed data at the output of the decoder:
\begin{equation}
    \mathrm{Loss}=\mathrm{MSE}\left(X,\widehat{X}\right),
\end{equation}
where $X$ is the input PD and $\widehat{X}$ is the reconstructed data. 
Thus, the AE uses an undercomplete representation to limit the amount of information that can flow through the network. By penalizing the network according to the reconstruction error, our model can learn the most important attributes of the input data and how to best reconstruct the original input from an encoded state. This learned encoding will ideally represent the latent attributes of the input data, thereby performing data compression. 

{\it Stage 3: Encoding and Quantization:}
A mesh node represents a $33\times 37$ particle histogram on ITER. The autoencoder model~(see Figure~\ref{fig:autoencoder}) reduces the $33 \times 37$ double precision floating point numbers to a bottleneck layer of the autoencoder. Based on empirical results, the size of the bottleneck layer can be as small as $4$. The reduced output coefficients for all images constitute the latent space. It is possible to reduce the data size for representing the latent space by further using quantization. We apply a product quantization (PQ)~\cite{jegou2010product} scheme. PQ is a vector quantization~\cite{vq} technique that breaks a high-dimensional vector into numerous lower-dimensional vectors and quantizes them independently using a simple clustering algorithm such as K-means. The decomposition to the high-dimensional vector and the quantization effectively create a Cartesian product of smaller subspaces to represent the original vector space (hence the term `product' quantizer). Although very simple, the advantage of this approach is that the number of vectors that can be represented accurately by the codebook can be pretty significant for the size of the codebook. The size of the quantized data depends on the number of bits that a PQ dictionary uses to represent codewords in the codebook (PQ bits) and the dimension of input vectors. Assume we have an $L$-dimensional vector consisting of floating-point numbers from an AE encoder and let $B$ be the PQ bit encoding. Then, each floating-point number can be represented by $B$ bits, and therefore, the size of the quantized vector is $L\times{B}$ bits.

{\it Stage 4: Residual Post-processing That Guarantees Error Bounds on Primary Data:}
The autoencoder model is helpful in accurately representing a large number of images (after reconstruction) within the user-defined error bounds.   However, for a small fraction of images, the model produces errors above the required bounds. For such images,  error guarantees are provided by computing the residuals and then using an existing state-of-the-art data compressor that provides error guarantees to compress the residuals. The residuals are the difference between the primary and reconstructed data computed by the AE. There are a variety of data compressors, such as MGARD~\cite{MGARD_1,MGARD_2,MGARD_3}, SZ~\cite{tao2017significantly,zhao2021optimizing} and ZFP~\cite{lindstrom2014fixed}. We focus on MGARD because MGARD is the state-of-the-art compression technique that guarantees error bounds on primary data and has a QoI-preservation step. 

MGARD achieves the guarantees using truncated basis function representations. The process recursively divides the input domain with a fixed basis function within each subdomain to represent the data. If the fixed basis cannot meet the error requirement, it is abandoned, and the subdomain is further divided. This way, error guarantees are met, and a subdivision procedure plays out per image. Depending on the methods of dividing grids, MGARD has two versions. One is MGARD uniform, which only focuses on guaranteeing PD error bounds and achieving the best possible compression without considering QoI preservation. It divides grids and assigns quantization bins uniformly. The other version of  MGARD  considers guaranteeing PD error bounds and the preservation of linear QoIs. It uses an additional $s$ parameter that assigns different weights and quantization bins depending on the QoIs. The second version requires the division of grids in a non-uniform fashion. MGARD non-uniform achieves lower compression than MGARD uniform with the same error bound because it uses precise quantization bins for some coefficients to preserve QoIs. Since our approach has a separate (and more effective) method for guaranteeing QoIs, we limit ourselves to the uniform version.

The use of MGARD for residual processing is limited to only the images for which AE cannot provide the user-defined error threshold. MGARD has a parameter that has to be chosen for guaranteeing the PD error. To efficiently choose this parameter value, we use a binary search on the parameter space to provide guarantees on the PD error.
\eatme{
 Furthermore, we use the reconstructed residuals selectively based on an error level of each instance. We need to store not only the compressed data from an AE, but the compressed residuals as well, which affects the overall compression ratio. Therefore, we set error thresholds to reduce the number of residuals being used. If the error of an image is larger than the threshold, we process the residual of the image with MGARD, so that the error doesn't exceed the threshold. 
}

\eatme{

This paper introduces a substep here to find the best error bound that is supplied to MGARD for processing the residuals. Of importance here is the fact that the error bound of $0.001$ applies to the image but not directly to the residuals. Our goal is to automatically determine the best error bound for the residuals that guarantees the PD error while providing the maximum compression. We empirically approach this problem using a binary search on the possible range of error bounds to determine the best outcome.
}

{\it Stage 5: Constrained Optimization to Preserve QoIs:}
The QoIs we seek to preserve in XGC are essentially moments and energy properties of the PD. We first set up the different moment constraints and then proceeded with a standard Lagrangian constraint satisfaction approach to QoI preservation. Because the loss function we use (distance between the reconstructed PD and the PD obtained after constraint satisfaction) is convex and the constraints linear, we employ a standard dual maximization approach using Newton optimization [because constraint cardinality is $\mathcal{O}(1)$]~\cite{jaemoon2}. We introduce a step-size parameter, $s$, to Eq. 26 in~\cite{jaemoon2} to control the convergence rate of the $\lambda$s, as shown in the equation below:
\begin{equation}\label{eq:newtstep}
{\boldsymbol{\lambda}^{k+1}=\boldsymbol{\lambda}^{k}-sH^{-1}\left(\boldsymbol{\lambda^{k}}\right)g\left(\boldsymbol{\lambda^{k}}\right),}
\end{equation}
The step-size parameter, $s$, is set to $1.0$ by default, and the maximum number of iterations is set to $50$. 

Thus, Stage 5, the following processing happens:
\begin{enumerate}
\item The compressed tensor is reconstructed using the AE decoder.
\item The true QoIs are computed or read for the constraint satisfaction step. These QoIs are computed from the original tensor.
\item Post-processing uses the reconstructed tensor and the true QoIs as input and computes the $\lambda$s. We compute four $\lambda$s per mesh node (one for each QoI). More about the post-processing step may be found in~\cite{hipc2022}.
\end{enumerate}

Data compression and post-processing are complete at the end of Stage 5. The data organization in the ADIOS buffer populated by each processor is organized as a header of length 44 bytes, the AE model weights, the quantized codes, the PQ table, the compressed residual bytes, and the Lagrange multipliers - the $\lambda$s. The header contains metadata about the compression scheme used, the byte boundaries of compressed data, and the total size of the data buffer. ADIOS collects the buffer from all processors and writes it out to disk. 

Data decompression using our methodology employs an ADIOS reader to read the data compressed and post-processed using our software pipeline, as described above. The first step in the decompression process locates the start and end bytes, stores the compressed tensor, and reconstructs the tensor using an appropriate decompression routine. The second step uses the reconstructed tensor and the supplementary information about the $\lambda$s and applies the Lagrange correction to each tensor value. The resulting tensor is the decompressed output.

\section{Experiments\label{sec:expt}}
This section presents the experimental setup and results of applying our compression pipeline to the ITER simulation data. Section~\ref{sec:setup} shows the various setup details, such as platform setup, a brief description of the dataset, the I/O mechanism, and the metrics and baseline used throughout for evaluating the results in terms of compression performance and accuracy. The following Sections~\ref{sec:aetraining} through \ref{sec:errors} describe experiments that explain the rationale behind setting up the soft parameters to the pipeline, leading up to the last Section~\ref{sec:iterexpt} that describes the compression results on all simulation timesteps of ITER. 
\subsection{Setup\label{sec:setup}}
\subsubsection{Platform\label{sec:platform}}
Summit, a supercomputer at Oak Ridge National Laboratory~\cite{hines}, is the platform for our experiments. Summit consists of 4608 IBM Power9 compute nodes, each with 42~cores and 6~Nvidia Volta GPUs. Each node has 1.6~TB of node-local NVME storage. The IBM Spectrum Scale GPFS parallel file system is used for large-scale I/O. This paper uses $32$~Summit nodes ($6$~processes per node, for a total of $192$~processes, and 1~GPU per process) to compress ITER data.
\subsubsection{ITER Dataset\label{sec:iter}}
We collected the outputs from an XGC simulation for the ITER geometry using 1024 Summit nodes (6144 GPUs) and focused on compressing $F$ data (the data for particle distribution function) over multiple steps. The XGC ITER run simulates 34 billion virtual electrons and ions. 
The simulator creates $F$ data where each particle type (electrons or ions) takes 81~GB per step, representing the kinetics of particle motions in five-dimensional physics (three dimensions in real space and two dimensions in velocity space). XGC saves $F$ data in the file system by forming a multidimensional array of the shape of $(8, 33, 1.1\mathsf{x}10^6, 37)$, representing the data for 8 poloidal planes, 1.1 million mesh points, and $33$-by-$37$ resolution velocity space. One single ITER timestep with ions and electrons completes a simulation timestep in about $1$ minute on $1024$ Summit nodes. Our objective (details to be provided later) is to achieve the compression in less than this time. Such fast compression times would help achieve a software pipeline where data generated from the previous step is compressed in parallel with the simulation of the current step.

\subsubsection{ADIOS}
ADIOS-2, the Adaptable Input Output System, is an I/O framework designed to manage scientific data at scale while providing scalable parallel I/O performance on HPC systems. 
ADIOS provides a unified application programming interface (API) with a level of abstraction suitable for how data are produced and consumed in scientific applications. In addition, ADIOS provides various data services to reduce the cost of integrating different data-related operations, such as metadata collection, transportation, and compression~\cite{godoy2020adios, logan2020extending}.

ADIOS supports a high-level operator mechanism, known as a callback, to allow a set of operations to perform on the data in memory before it is written or saved in a specific format. With this operator mechanism, ADIOS can easily support several different lossless and lossy compression methods. In addition, ADIOS provides a systematic way to extend the capability with a user-defined compression operator where users can define their data operations and compression. We focus on using the ADIOS callback mechanism to develop a compression method with post-processing for XGC data.

\subsubsection{Metrics\label{sec:accu}}
We use the normalized root mean square error (NRMSE) to evaluate the compression quality and the compression ratio to measure data size reduction. NRMSE is a relative error used to measure both the primary data and the QoI errors. NRMSE for primary data is defined as follows: 
\begin{equation}
\text{NRMSE}(u, f) = \frac{\sqrt{\sum_{i=0}^N(u_i-{f}_i)^2/N}}{\max(u) - \min(u)},\label{eq:nrmse}
\end{equation}
where $u$ is the original data, ${f}$ is the reconstructed data, and $N$ is the number of degrees of freedom in $u$. When Eq.~(\ref{eq:nrmse}) is used for computing QoI errors, $u$ is the QoI computed on the original data, and ${f}$ is the QoI computed using the reconstructed data. Our specified bounds for PD error is $0.001$ and for QoIs $0.0001$. Our method's QoI errors are significantly lower than the specified bound. Eq.~(\ref{eq:nrmse}) is also used to compute image NRMSE, where $N$ is the product of the cardinality of velocity coordinates $V_x$ and $V_y$, which for ITER is $33 \times 37$ = $1221$.

The compression ratio is the ratio of the size of the original data to that of the compressed data. A good proxy for the level of compression that can be achieved using our pipelines is based on the fraction of images for which AE can provide user-defined primary data error bounds. Effectively for these images, MGARD (or other methods that guarantee PD error bounds) is not required for residual processing.
\begin{figure}[tbp]
    \centering
    \subfloat[Errors in primary data (PD)]
    {\includegraphics[width=\columnwidth]{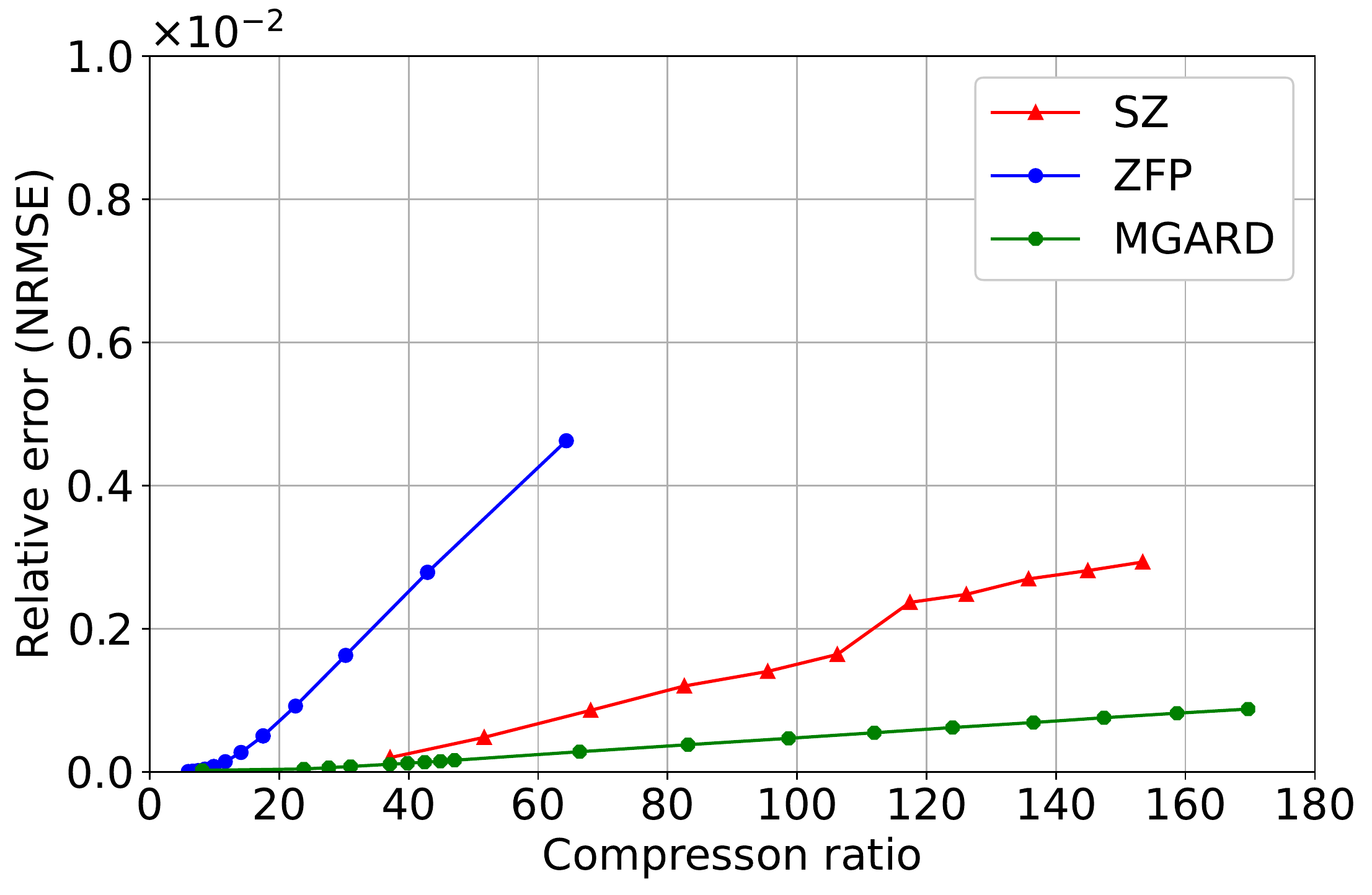}} \quad
    \subfloat[Errors in quantities of interest (QoI)] 
    {\includegraphics[width=\columnwidth]{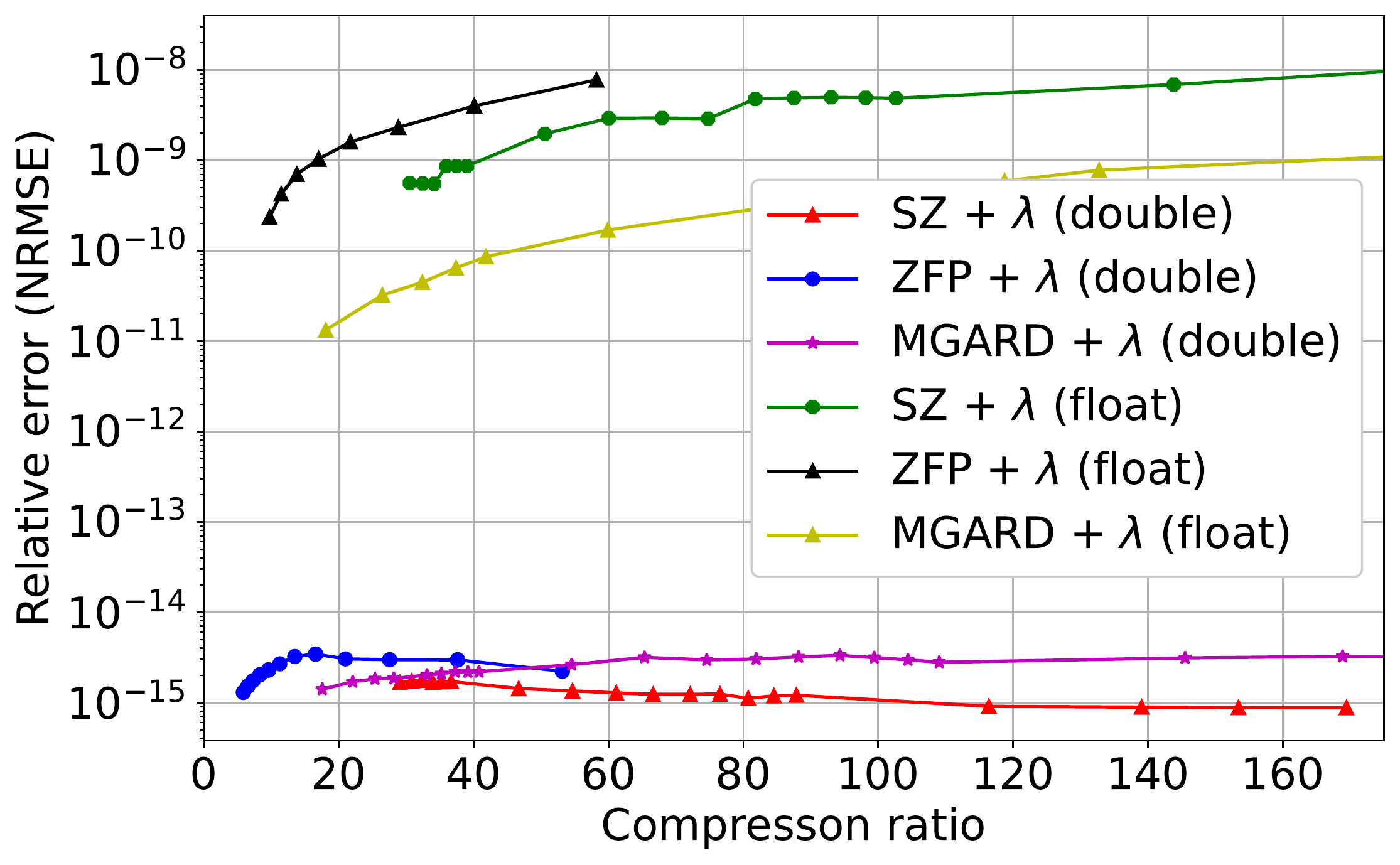}}
    \caption{(a) Comparison of NRMSE errors on primary data for  MGARD, SZ, and ZFP, separately based on the compression. These results show that MGARD is better than the other two methods for XGC. (b) Errors on QoIs based on  using the lossy compressed data from MGARD, SZ, and ZFP and then applying a post-processing step. The maximum error of the four QoIs is shown on the ordinate axis. For all three compressors, the NRMSE errors of QoIs computed from the reduced PD drop down to machine epsilon (when double precision is used for $\boldsymbol{\lambda}$) in post-processing.}
    \label{fig:combo1}
\end{figure}
\begin{figure}[tbp]
    \centering
    \subfloat[Distributed Training]
    {\includegraphics[width=0.95\columnwidth]{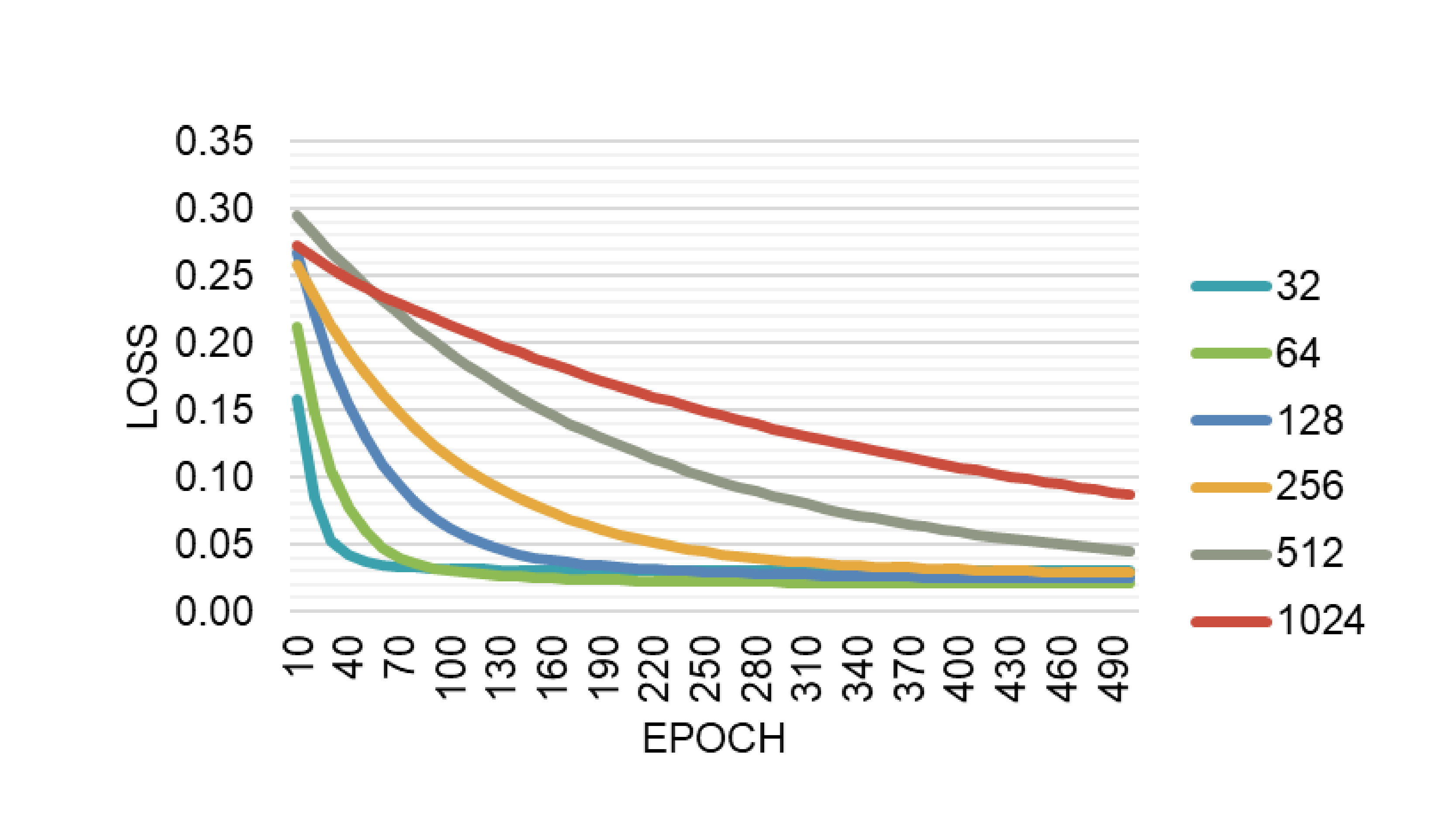}}
    \quad
    \subfloat[Centralized Training]
    {\includegraphics[width=0.95\columnwidth]{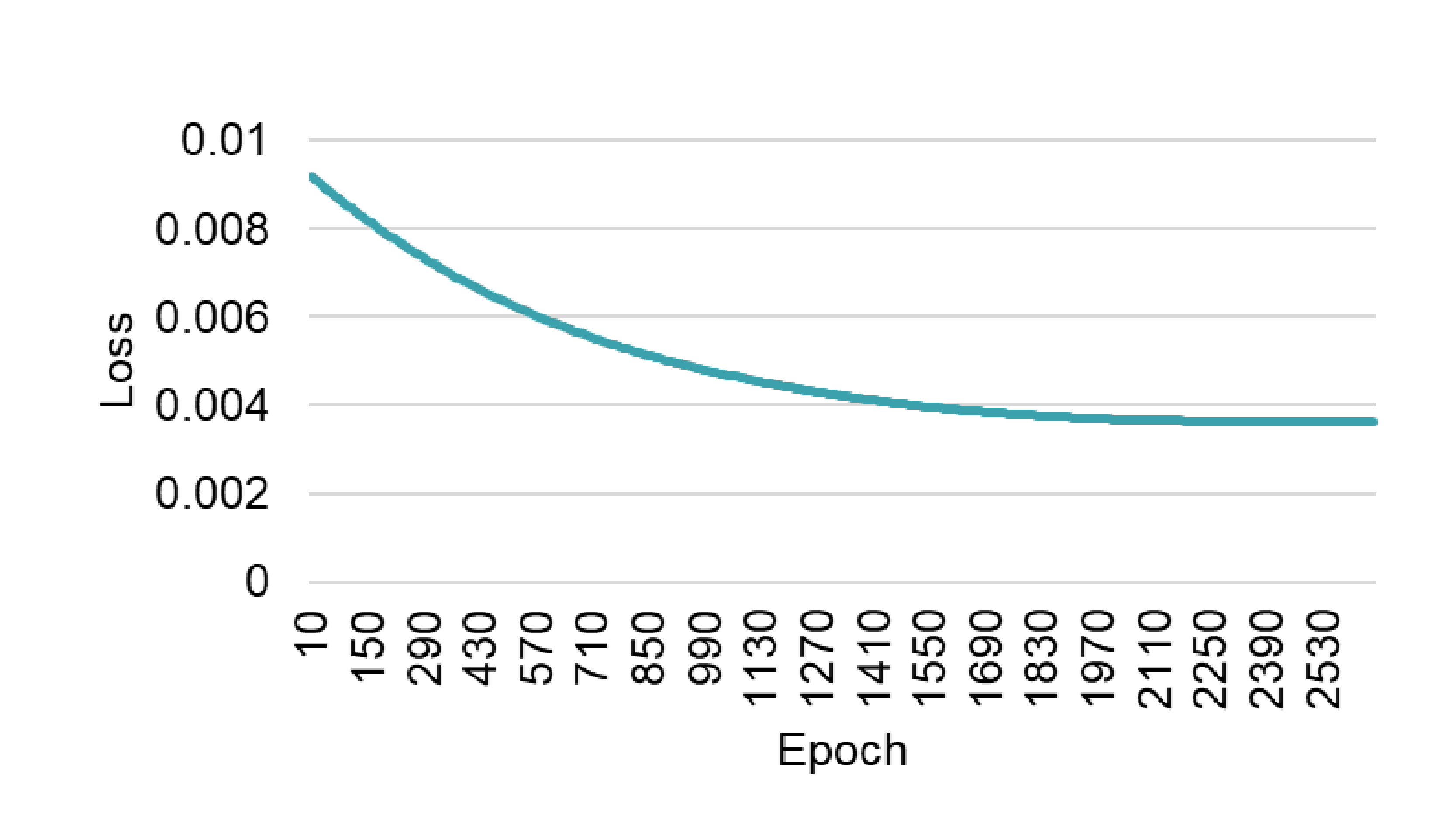}}
    \caption{(a) Loss convergence using distributed training for different batch sizes in distributed training on the 192nd processor. Smallest batch size 32 converges the fastest, as expected. (b) Loss convergence using centralized training takes about 1800 epochs on 32 Summit nodes using a batch size of $32$; the total runtime is $40$ minutes. Incremental training was used for this experiment on simulation timestep 220 and 2 epochs. The learning rate was $10^{-5}$, and the converged loss value was about $0.0036$ in both cases.}
    \label{fig:centralized}
\end{figure}
\begin{figure}[tbp]
    \centering
    \includegraphics[width=0.75\textwidth,trim={5cm 0cm 0cm 0cm},clip]{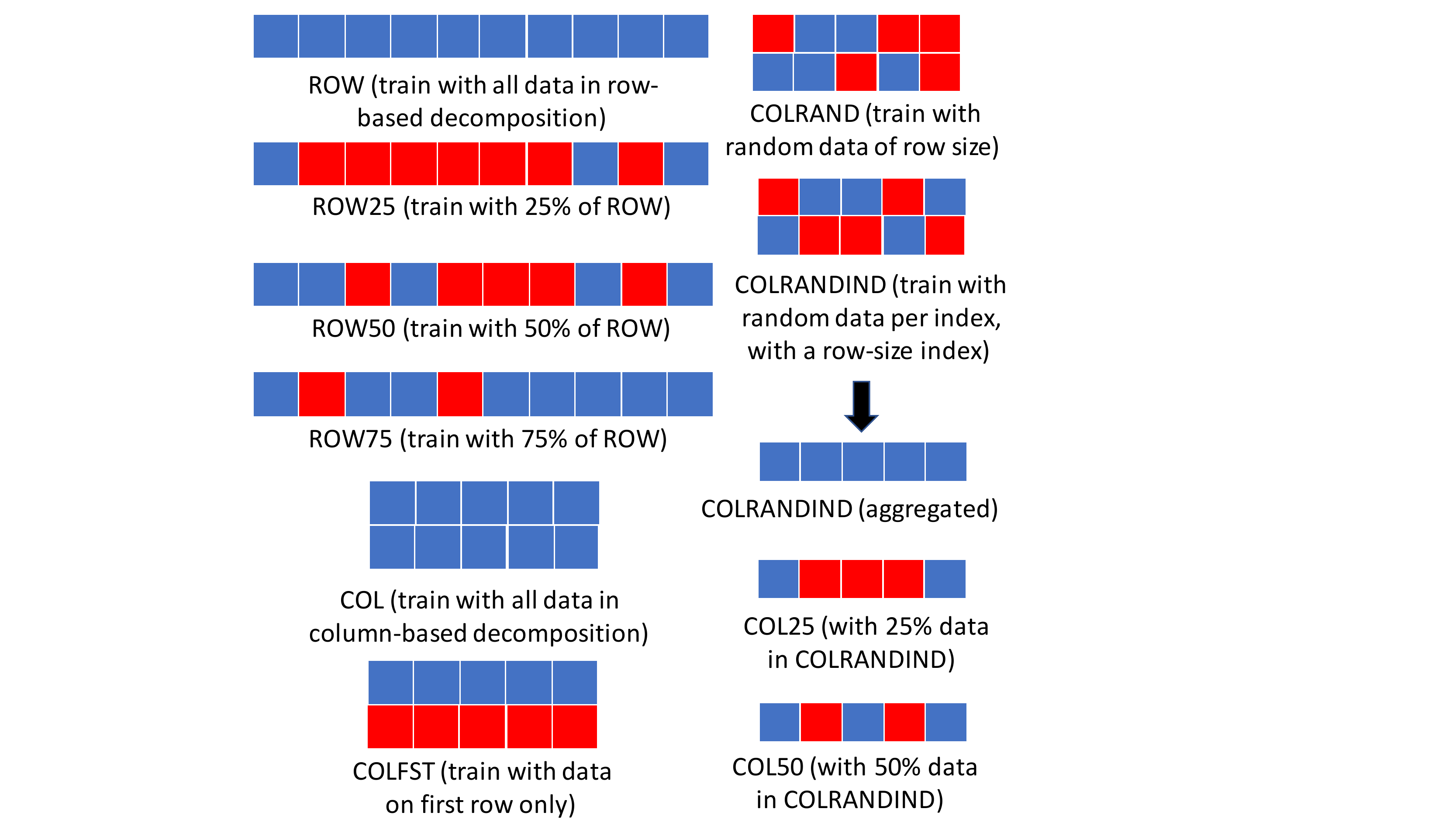}
    \caption{Data selection in a processor for AE training for row-based and column-based decomposition. The blue squares represent data used for training, whereas the red squares represent data that are dropped. For example, in COLRANDIND, data for each index 1, ..., 5 are chosen randomly from either the first or the second row.}
    \label{fig:tp}
\end{figure}
\begin{figure*}[ht]
    \centering
    \subfloat[AE accuracy for full training with different training-data selection schemes]
    {\includegraphics[width=2\columnwidth,trim={0cm 5cm 0cm 4cm},clip]{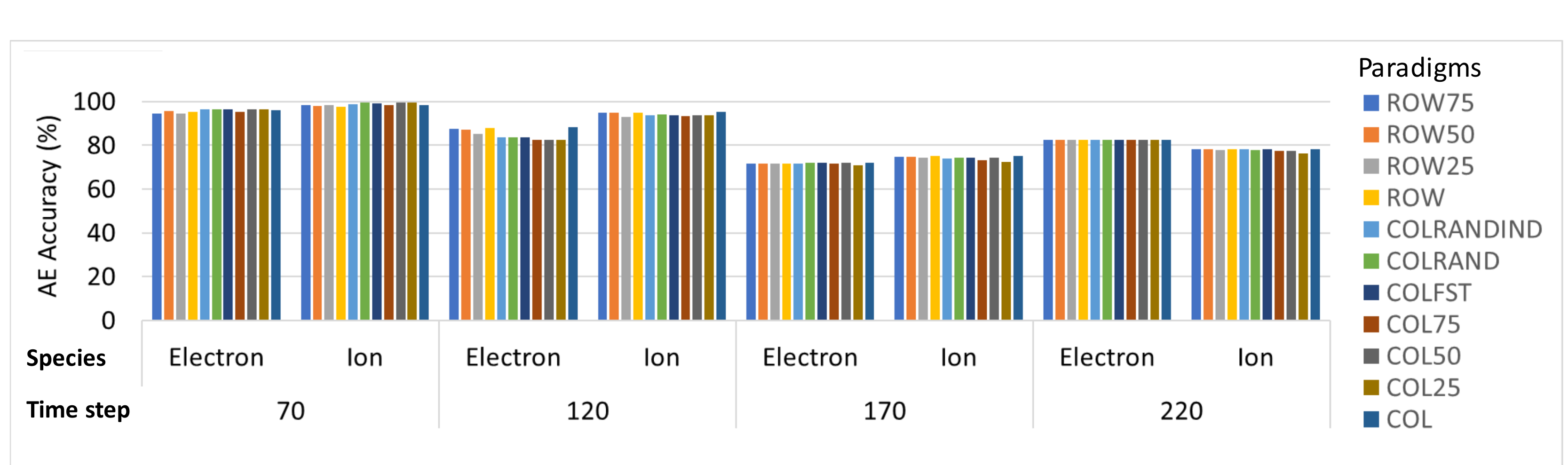}}
    \quad
    \subfloat[Training time in seconds for full training with different training-data selection schemes]
    {\includegraphics[width=2\columnwidth,trim={0cm 4cm 0cm 4cm},clip]{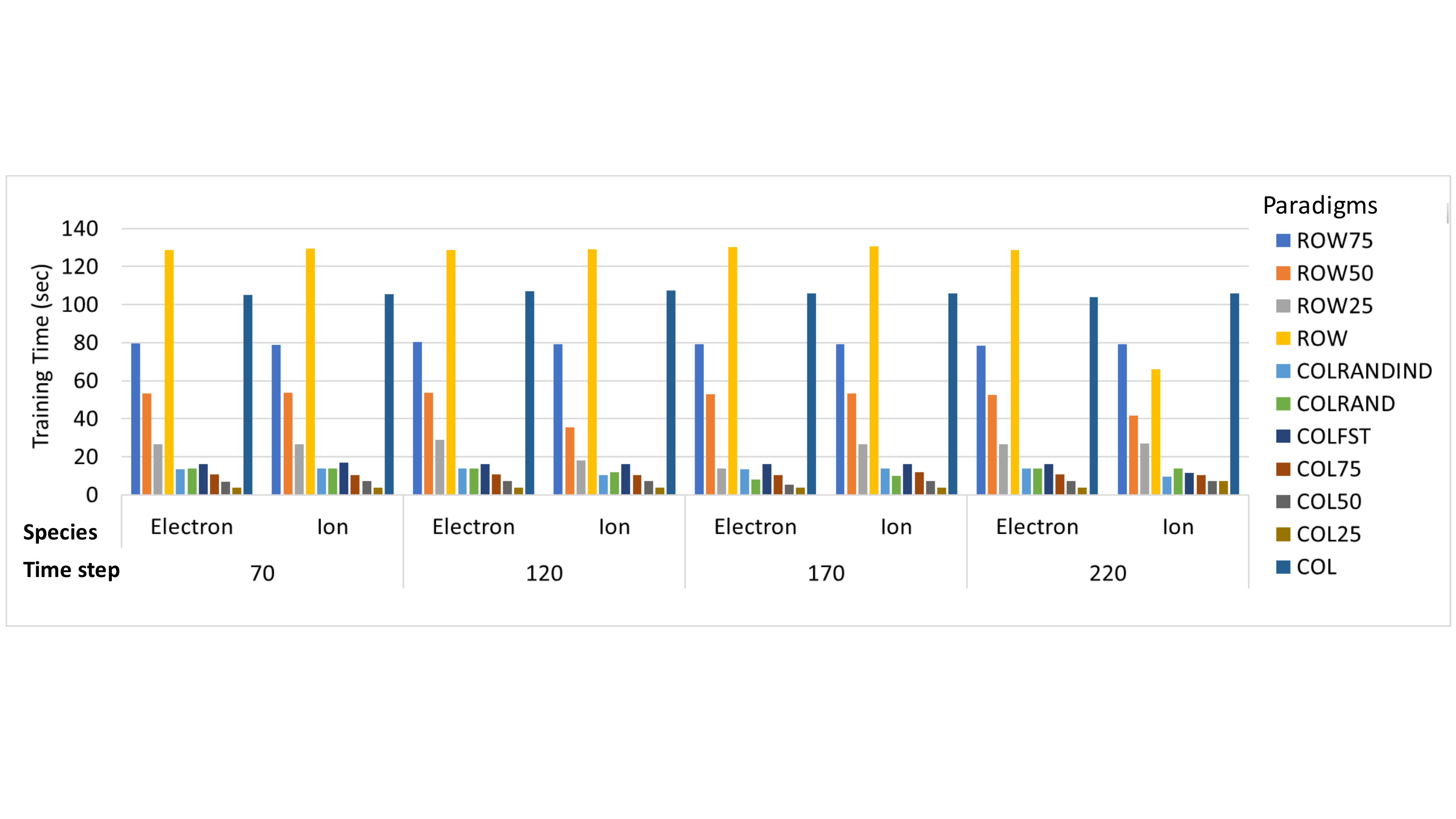}}
    \caption{(a) The AE accuracy among the various data selection schemes is comparable for each particle type and each time step. The accuracy is evaluated as the percentage of images decoded accurately by the AE and not requiring residual processing. (b) Training time varies based on the amount of data used for the training. ROW and COL use the entire dataset, whereas ROW25, ROW50, ROW 75 use respectively 25, 50, and 75\% of the dataset. COLFST, COLRAND, COLRANDIND uses 12.5\% of the dataset (because there are $8$ rows of data in our test; so each row used in the selection scheme is 12.5\% of the data). The remaining paradigms COL25, COL50, and COL75 use 3.125, 6.25, and 9.375\% of the COLRANDIND data, respectively.  Overall from (a) and (b), we select the COLRANDIND data selection scheme, which takes about 13.7 and 11.8 seconds, respectively, to train a model for electrons and ions from scratch.}
    \label{fig:aet}
\end{figure*}
\begin{figure}
    \centering
    \includegraphics[width=\columnwidth,trim={0cm 2cm 0cm 2cm},clip]{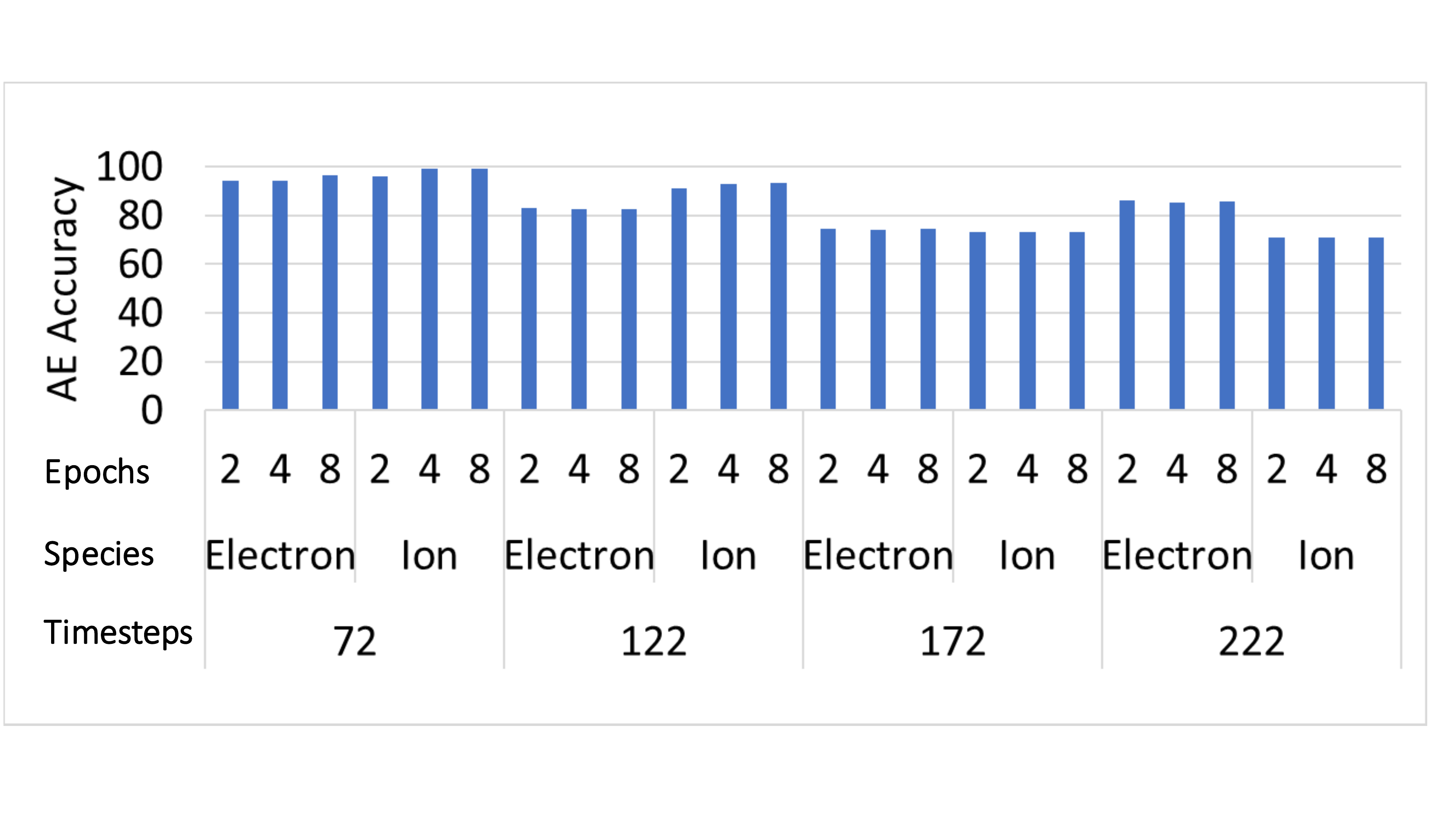}
    \caption{Incremental training using $2$, $4$, and $8$ epochs with the COLRANDIND data selection scheme. The average accuracy is $84.71\%$, $84.78\%$, and $84.82\%$ for epochs $2$, $4$, and $8$, respectively. Using full training gives us an accuracy of $84.97\%$}
    \label{fig:epoch}
\end{figure}
\subsection{Lossy Compressors MGARD, SZ, and ZFP}
Figure~\ref{fig:combo1} plots the NRMSE of PD and four QoIs using the lossy compressed data produced by MGARD, SZ, and ZFP. In Figure~\ref{fig:combo1}(a), we show the NRMSE error of PD when the data are reduced without post-processing. In Figure~\ref{fig:combo1}(b), we show the maximum NRMSE errors measured among four QoIs which are computed from the lossy compressed and post-processing updated PD. As the figure illustrates, the post-processing pushes the errors of QoIs into machine epsilon ranges, regardless of which compressors are used. Our results also show that the PD and QoI errors are lower in data compressed with MGARD than those with the other two compressors. Among those target error ranges and by L2 error metric, MGARD delivers larger compression ratios than SZ and ZFP~\cite{gong2021maintaining}. Therefore, in the rest of this section, we only show the results that pertain to MGARD.

We evaluate the performance of our compression pipeline with and without AE (referred to as the non-AE compression pipeline introduced in~\cite{hipc2022}). Both our AE and non-AE pipelines use MGARD to guarantee PD error bounds. 
\eatme{
Although other primary data compressors, such as SZ and ZFP, may be used in our data compression pipeline, we found MGARD to be superior because it has lower PD errors than SZ and ZFP for the same level of compression.
}
\subsection{AE Implementation Details}
We implemented an autoencoder (AE) using PyTorch C++. The network architecture consists of one fully connected layer for the encoder and the decoder with linear activation. The model weights at the encoder are transposed at the decoder. So we have a single set of model weights, which improves the stability of the AE and reduces the model size. We used the Adam optimizer \cite{kingma2014adam} with a learning rate of $0.001$ and set the batch size to be $128$.

\subsection{AE Training\label{sec:aetraining}}
This section presents AE training methods and results. \eatme{The data selection methods are different based on the domain decomposition technique.} The two training paradigms introduced here are full training and incremental training. Full training involves training the neural network model from scratch, starting with zero weights. Thus, the term `full training' is synonymous with `training from scratch' and used interchangeably in the paper. Incremental training begins with a previously trained model on another timestep and updates the weights based on the data in the current timestep.

\subsubsection{Centralized versus Distributed Training}
We experimented with both the centralized and distributed training paradigms. The centralized training was implemented using distributed data parallel (DDP) functionality of PyTorch that enables one to train a model across multiple machines using data parallelism. Multiple MPI processes are spawned, and a single DDP instance is created for every process. DDP synchronizes gradients and buffers using a collective communication scheme that may be found in the torch.distributed package. DDP performs gradient synchronization across processes~\cite{ddp} to build a centralized model. On the other hand, each processor performs training independently in the distributed training, resulting in 192 models for the 192 processes (from Section~\ref{sec:platform}, the compression runs on 32 Summit nodes, hence there are 192 processes).

We found in our experiments that the centralized training takes about $1800$ epochs to converge to an acceptable mean squared error (MSE) loss compared to $100$ epochs for the distributed training when the AE model is trained from scratch. Thus, the centralized training took about $40$ minutes to finish, whereas the distributed training took about $130$ seconds. Hence, we use distributed training and preset the epochs and batch size to $100$ and $128$, respectively.
The size of the weights used in the AE network for ITER data is image size ($33 \times 37$) times the size of the bottleneck ($4$), times the size of the floating point weights in bytes ($4$), which is $33 \times 37 \times 4 \times 4 = 19,536$ bytes. There are $192$ models in distributed training (each processor trains its model). Thus, there is a storage overhead of the model weights. Given the size of the dataset being compressed, this overhead is minimal.

\subsubsection{Full Training\label{sec:training}}
When we train the autoencoder model, one key question is whether we use the entire data available to a processor to train the model in every epoch. If not, what strategy could we use to select data for training? As one might expect, this strategy would depend on the domain decomposition. When we use row-wise decomposition, each processor has a portion of mesh nodes in a single row. The data selection  strategies we experimented with in this case are:
\begin{itemize}
    \item ROW: All data in a row-based domain decomposition
    \item ROW25: 25\% of ROW chosen randomly
    \item ROW50: 50\% of ROW chosen randomly
    \item ROW75: 75\% of ROW chosen randomly
\end{itemize}
On the other hand, when we use column-wise decomposition, each processor has a column of mesh nodes spanning over all planes. It turns out that for XGC data, the particle histograms are very similar on two mesh nodes in the exact relative location on two different planes. We utilized this similarity to select our training data and experimented with the following strategies:
\begin{itemize}
    \item COL: All data in a column-based domain decomposition
    \item COLFST: Data from the first row only, which exploits data similarity across rows in a column-based decomposition
    \item COLRAND: Data is chosen randomly for training, and data size is the same as a row size.
    \item COLRANDIND: Data chosen for training is row-sized, and data at each index is chosen randomly from data in all rows at that index.
    \item COL25: 25\% of COLRANDIND chosen randomly
    \item COL50: 50\% of COLRANDIND chosen randomly
    \item COL75: 75\% of COLRANDIND chosen randomly
\end{itemize}
Figure~\ref{fig:tp} illustrates the subsets of data considered for training for the row-based and column-based decomposition techniques.

Figure~\ref{fig:aet} shows a comparison of the different data selection schemes for distributed training for four different timesteps for electrons and ions. The timesteps $70$ and $120$ are from a less turbulent regime than the timesteps $170$ and $220$. Each experiment was repeated $5$ times, and the average training time and accuracy numbers are reported here. Figure~\ref{fig:aet} illustrates that the training time drops significantly for comparable accuracy when we exploited the similarity of data across different rows for a column-based decomposition and train on a row's worth of data. Given the accuracy versus training time profile, we choose the scheme COLRANDIND to perform the rest of the experiments. COLRANDIND takes about $13.7$ and $11.8$ seconds to train a model for electrons and ions, respectively. The particle histograms have a wide range of values [$2187.26$, $6.5719 \times 10^{17}$]. Hence, we normalize the data using Z-score normalization before training. We use a random data sampler for the training data loader. 

\subsubsection{Incremental Training}
During incremental training, we start with a model from a previous timestep, then train that model with data from the current timestep. The earlier timestep model can be already trained either from scratch or using incremental training from a prior timestep. Usually, just training for a few epochs is enough for incremental training. So, we compare the accuracy for training using $2$, $4$, and $8$ epochs for electrons and ions, as depicted in Figure~\ref{fig:epoch}. Fully trained models from timesteps $70$, $120$, $170$, and $220$ were used to incrementally train models for timesteps $72$, $122$, $172$, and $222$, respectively. The average time for training $2$, $4$, and $8$ epochs across all particle types and the four timesteps are $0.74$, $1.0$, and $1.51$ seconds, respectively. Because the accuracy is comparable, we train for $2$ epochs while using incremental training.

\subsection{Setting Up PQ Table Size}
A trained AE model encodes an image ($N=33\times37=1221$ floating point numbers) using a small dimensional vector. Our experiments with $4$, $8$, and $16$ dimensions showed that a $4$-dimensional latent vector gave as good a result as $8$ and $16$ dimensions. So, we set the latent space size to be four-dimensional. We chose a product quantizer (PQ) to quantize the latent space constituting all latent vectors. PQ uses the K-means algorithm under the hood to cluster the data column-wise and come up with $k$ clusters for each of the $4$ columns (dimensions) of the latent space. The number of clusters, $k$, must be specified beforehand, and we experimented with $k$=$16$, $64$, and $256$. Each of the $k$ clusters is represented using a centroid, and each centroid is mapped to a $log_2(k)$ bit index representing the cluster number. Thus, corresponding to our selection of $k$, the index is written using $4$, $6$, and $8$ bits. The centroids are mapped to their indexes (or cluster numbers) in a PQ table; hence, the number of entries in the PQ table is $k$ times the dimension of the latent vector space, and the PQ table size is $16k$ bytes for storing the floating point centroids (each $4$ bytes). Each processor executes PQ independently on its local data. We studied AE accuracy for the ions and electrons for the four different timesteps. We found that the average accuracy across particle types and timesteps is $85.07$\%, $85.59$\%, and $85.65$\%, respectively, for using $4$, $6$, and $8$ PQ bits. Because the accuracy is comparable, we use PQ=$4$ bits for quantizing the latent space in our experiments.
\begin{figure*}[tbp]
    \centering
    \subfloat[Errors in primary data (PD) for ions]
    {\includegraphics[width=0.95\columnwidth]{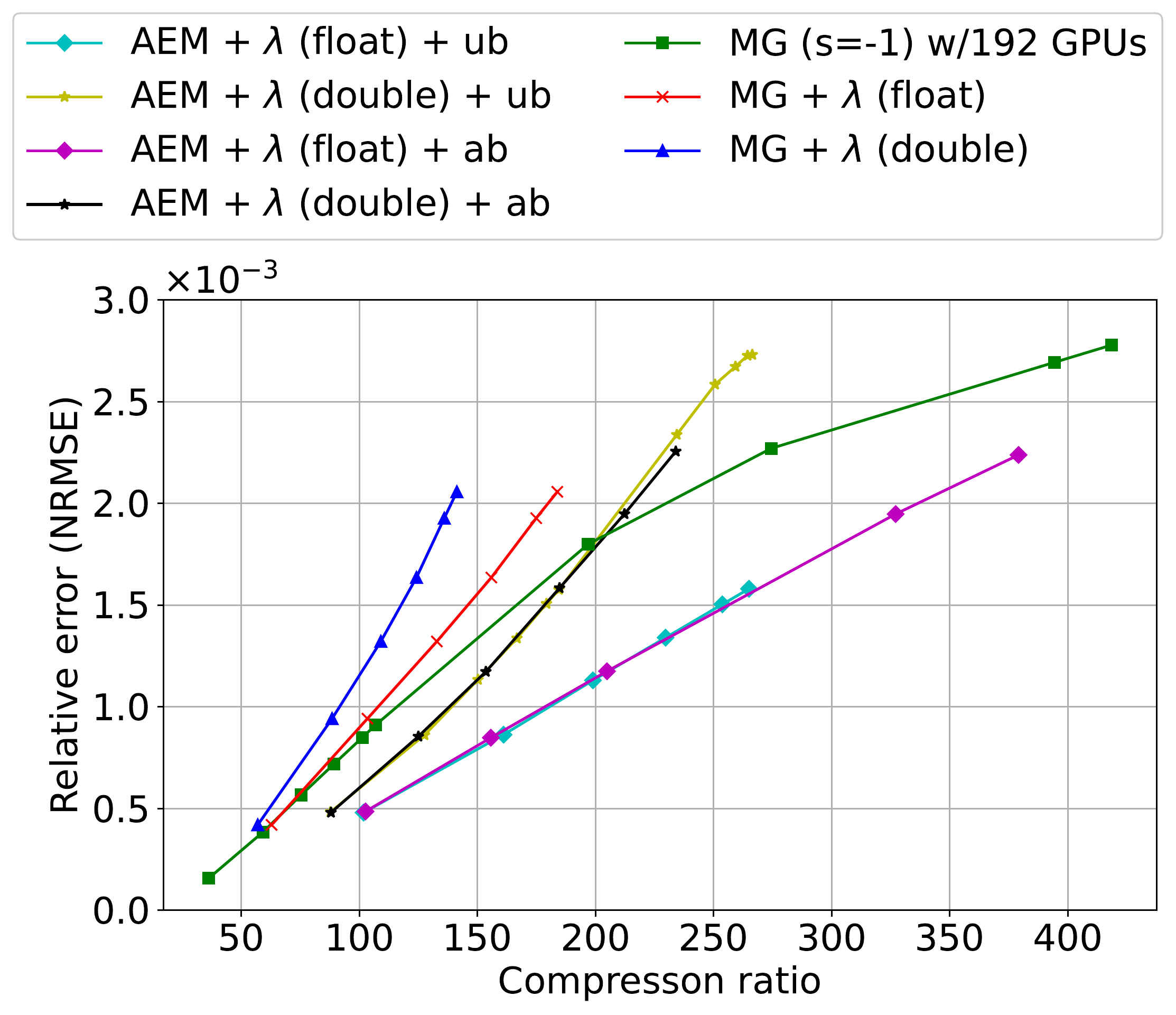}}
    \quad
    \subfloat[Errors in quantities of interest (QoI) for ions]
    {\includegraphics[width=0.95\columnwidth]{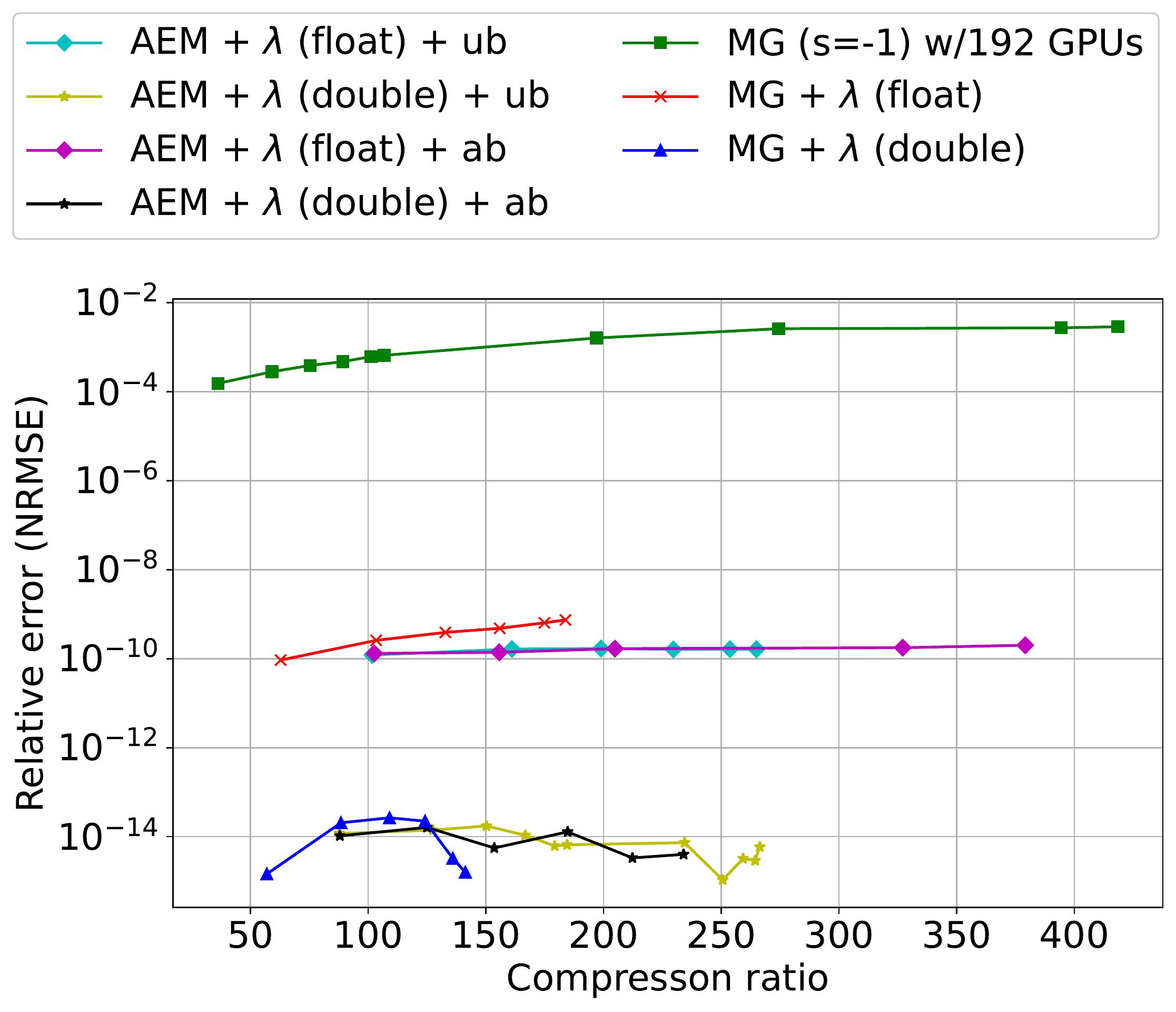}} \\
    \subfloat[Errors in primary data (PD) for electrons]
    {\includegraphics[width=0.95\columnwidth]{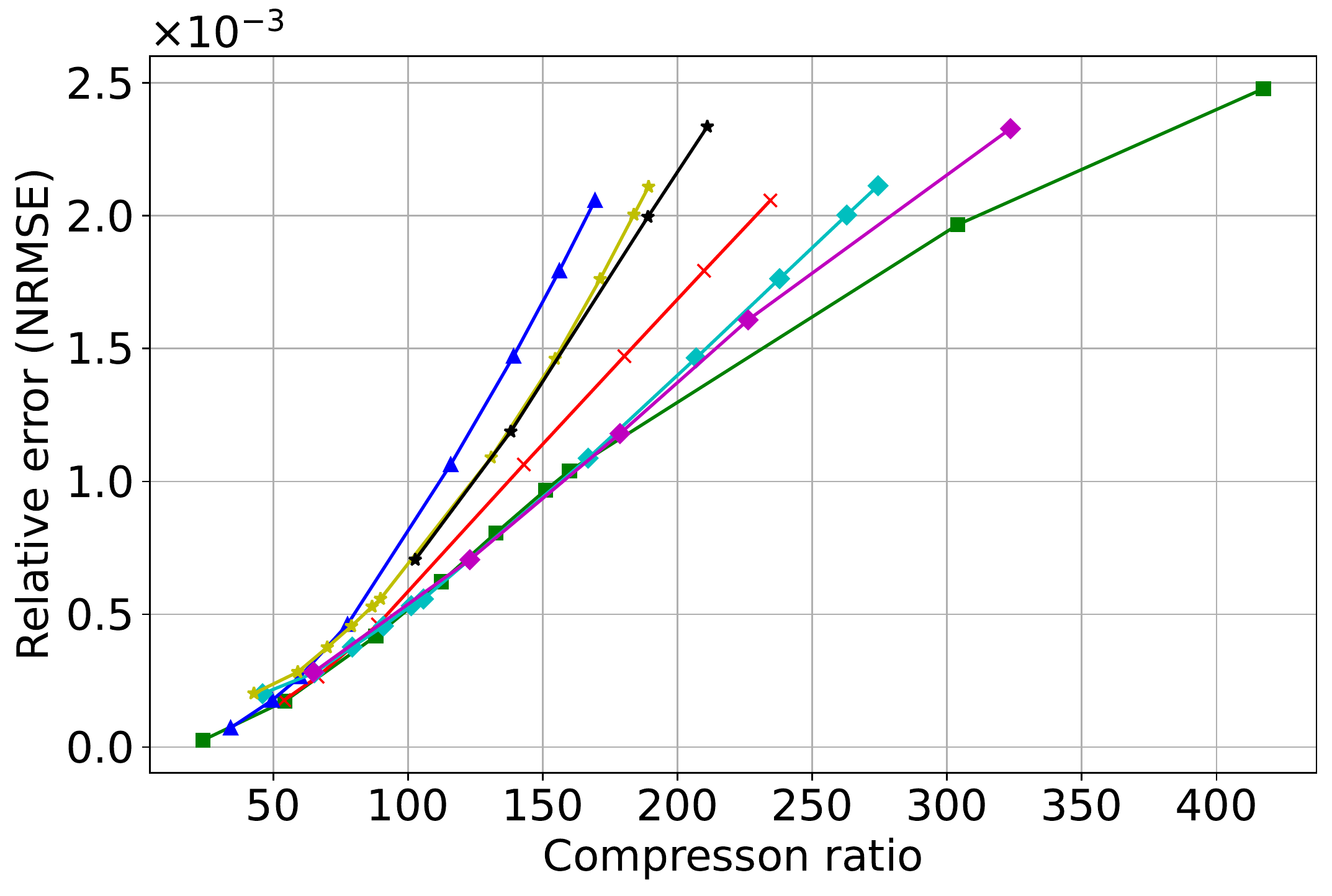}}
    \quad
    \subfloat[Errors in quantities of interest (QoI) for electrons]
    {\includegraphics[width=0.95\columnwidth]{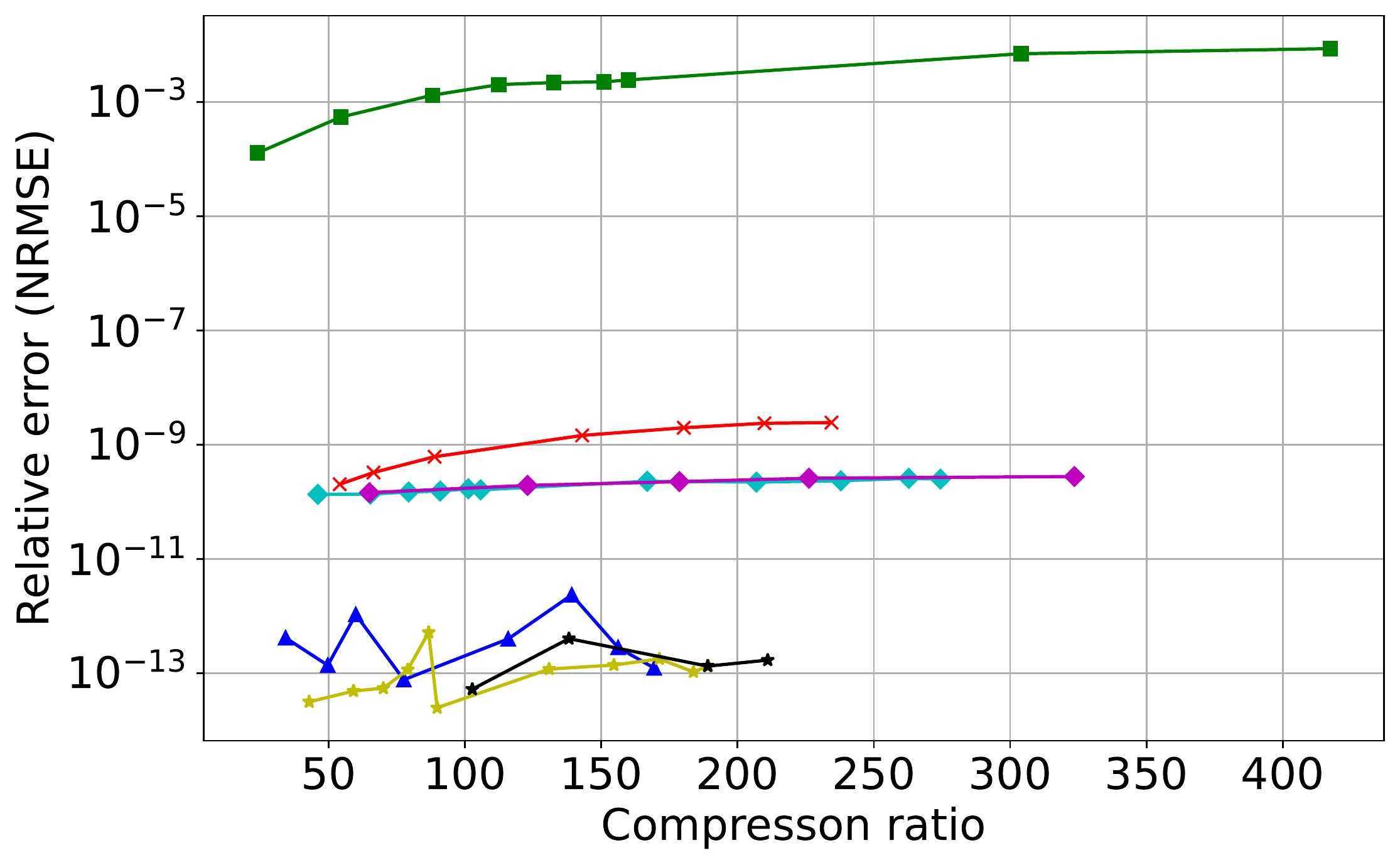}}
    \caption{Compression ratio versus error trends for ions and electrons for simulation timestep $170$. The primary data are lossy compressed using AE (with MGARD for compressing residuals) and using MGARD. The Lagrange parameters (i.e., $\boldsymbol{\lambda}$) are stored in their original (i.e., double) and a quantized precision (i.e., float). QoI errors are shown before and after post-processing. The maximum error measured among four QoIs is shown on the ordinate. We show that by applying post-processing, the errors of PD remain relatively constant, while the errors in the QoIs drop down to very low values.}
    \label{fig:ec}
\end{figure*}
\begin{table}
    \centering
    \caption{The three AE training paradigms that are used for compressing data across ITER timesteps.}
    \includegraphics[width=\columnwidth,trim={0cm 2cm 0cm 2cm},clip]{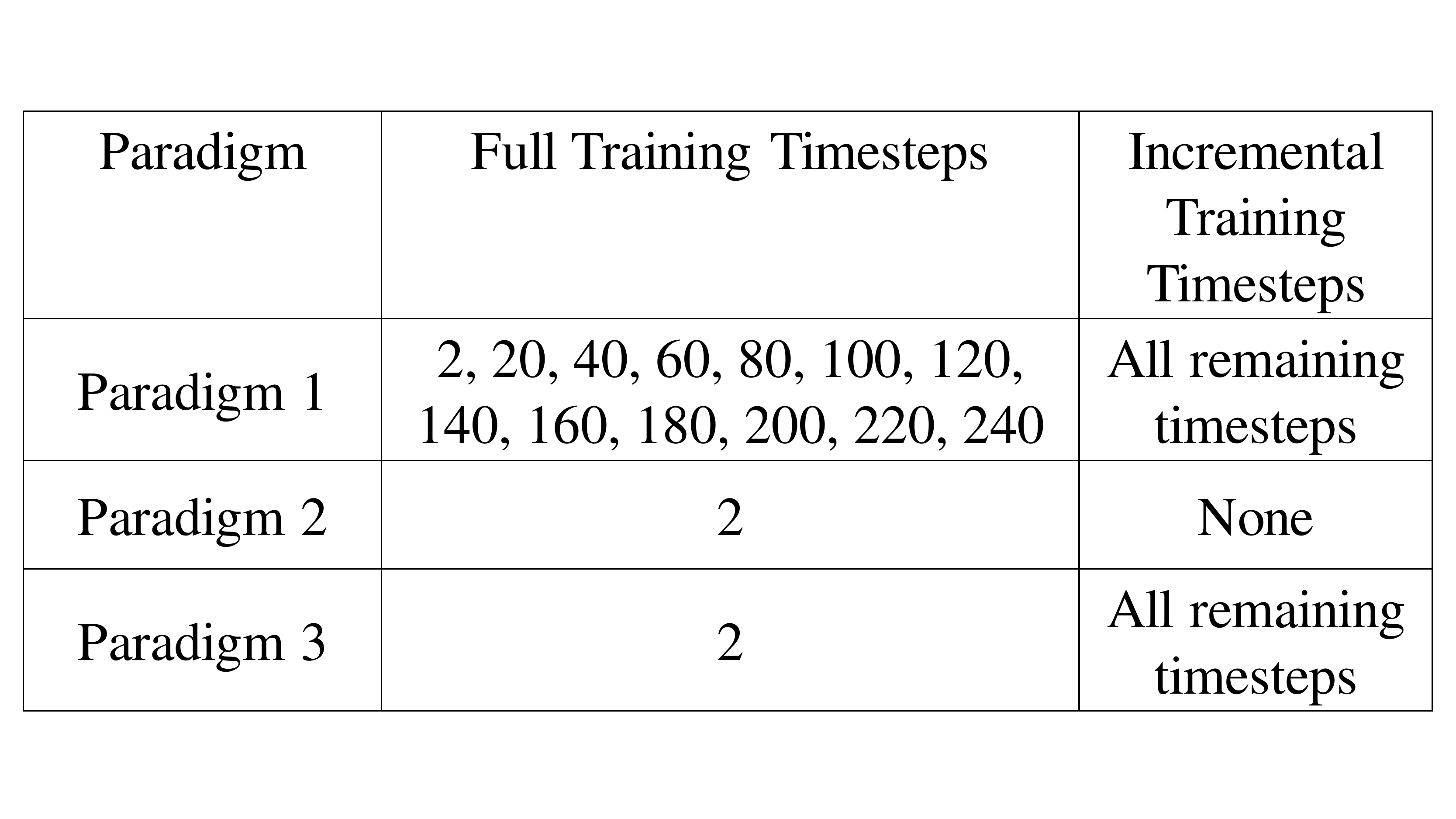}
    \label{tab:par}
\end{table}
\begin{table}
    \centering
    \caption{Compression ratios for the three AE training paradigms are comparable for ions. The results with electrons are similar.}
    \includegraphics[width=\columnwidth,trim={5.5cm 0cm 5cm 0cm},clip]{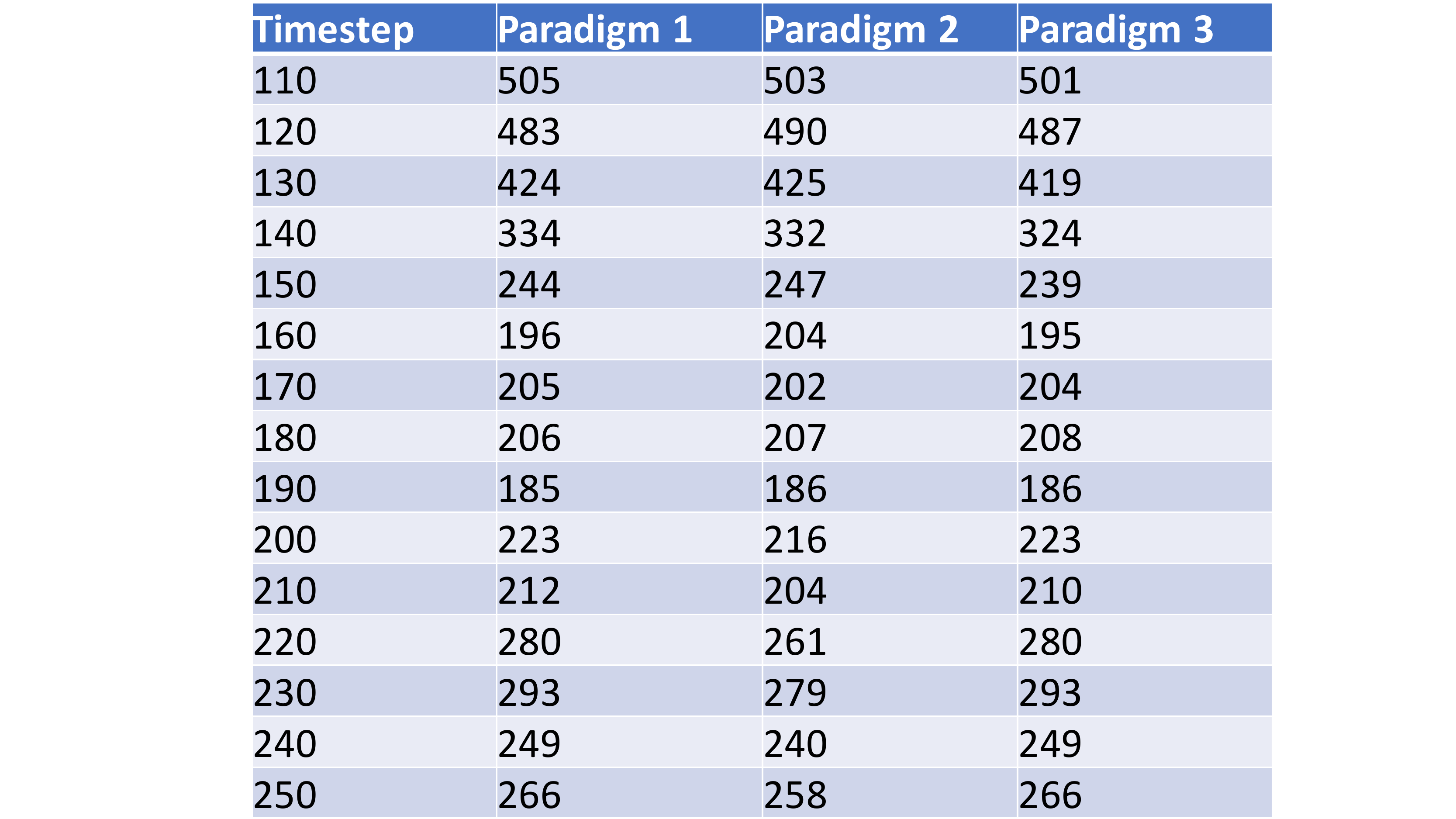}
    \label{tab:paradigmi}
\end{table}
\subsection{Error Trends with Compression Ratio\label{sec:errors}}
Figure~\ref{fig:ec} illustrates the PD and QoI errors versus the compression ratio trends for ions and electrons for ITER simulation timestep $170$, which is one of the timesteps for which we get a close to minimum compression. The average compression levels are much higher than that achieved in this timestep.

The various compression pipeline settings in the trend evaluation are as follows: 
\begin{enumerate}
    \item AEM + $\lambda$ (float) + ub: AE compressor with MGARD ($s$=$0$) for residual processing. Post-processing was applied and the $\lambda$s were stored as single precision floating point numbers. The abbreviation ub stands for user-specified error bound. The user must specify the error bound to MGARD for compressing the residuals.
    \item AEM + $\lambda$ (double) + ub: This setup is (1) above with the stored $\lambda$s having double precision accuracy.
    \item AEM + $\lambda$ (float) + ab: This setup is (1) above with the MGARD error bound for residual processing automatically determined using a binary search based on the PD error produced. The abbreviation ab stands for automatic error-bound determination. 
    \item AEM + $\lambda$ (double) + ab: This setup is (3) above with double precision $\lambda$s
    \item MG (s=-1) with 192 GPUs: This setup runs MGARD with $s$=$-1$ on $32$ Summit nodes ($192$ GPUs). The number of processors used is consistent with the rest of the experiments.
    \item MG + $\lambda$ (float): This setup adds the post-processing step to MGARD ($s$=$0$), and the $\lambda$s are stored as single precision floating point numbers.
    \item MG + $\lambda$ (double): This setup is the same as (6) above with double precision $\lambda$s.
\end{enumerate}
The settings (5) and (6) above pertain to the MGARD pipeline~\cite{hipc2022}. On the other hand, (7) is a standalone MGARD application with the $s$=$-1$ that optimizes for QoI errors. Compression pipeline setting (7) is different from the corresponding setup in~\cite{hipc2022} where $s$=$0$ was used. 

Figure~\ref{fig:ec} shows that QoIs computed on post-processed reconstructed data errors drop to as low as $10^{-12}$ when the $\lambda$s are stored in double precision and as low as $10^{-8}$ when the $\lambda$s are stored in single precision). Furthermore, the QoI error does not build up significantly with the compression ratio. 
\begin{figure*}[ht]
    \centering
    \subfloat[Compression ratios]
    {\includegraphics[width=0.9\columnwidth]{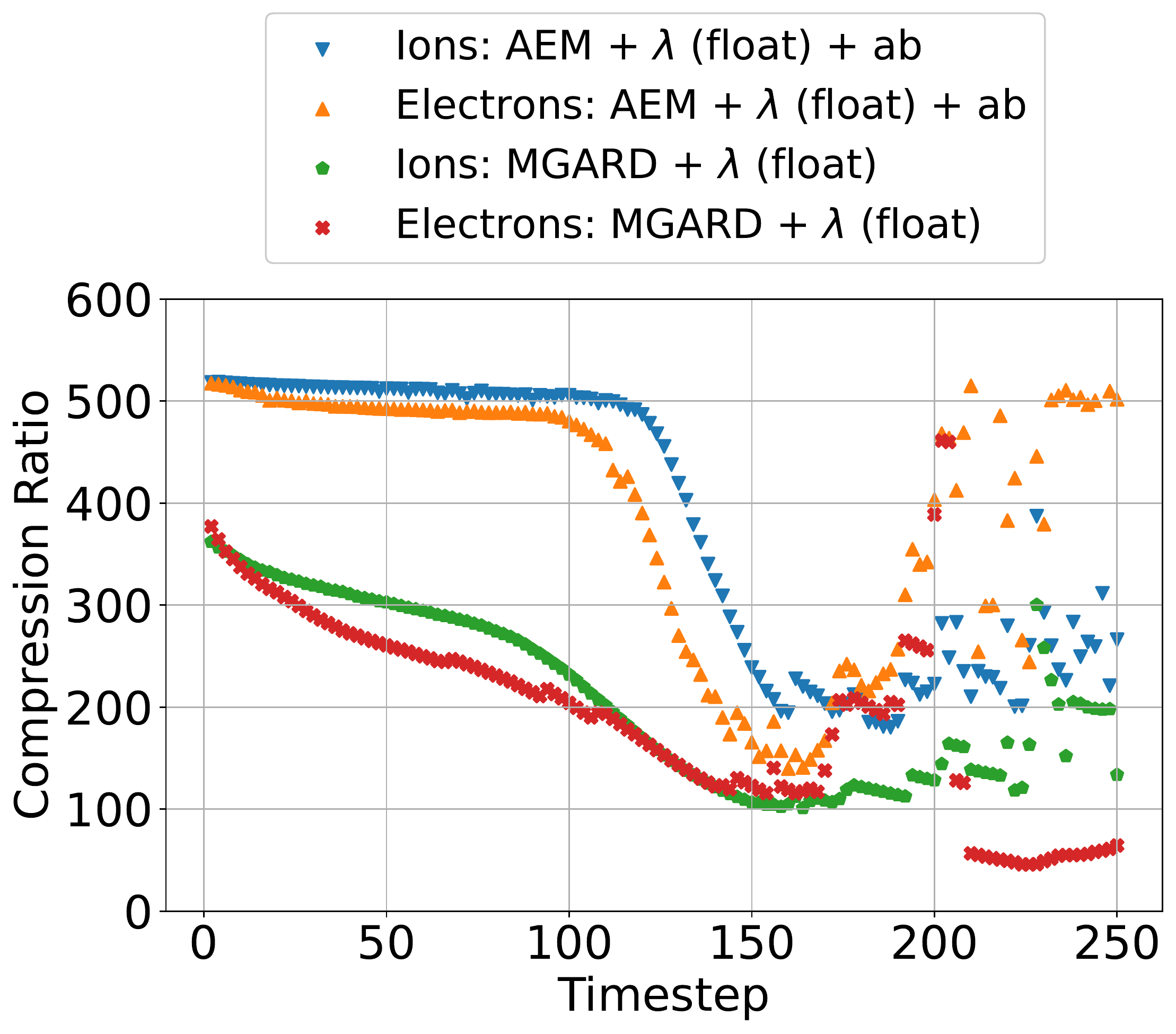}}
    \quad
    \subfloat[Ratio of compression using non-AE and AE pipeline]
    {\includegraphics[width=0.9\columnwidth]{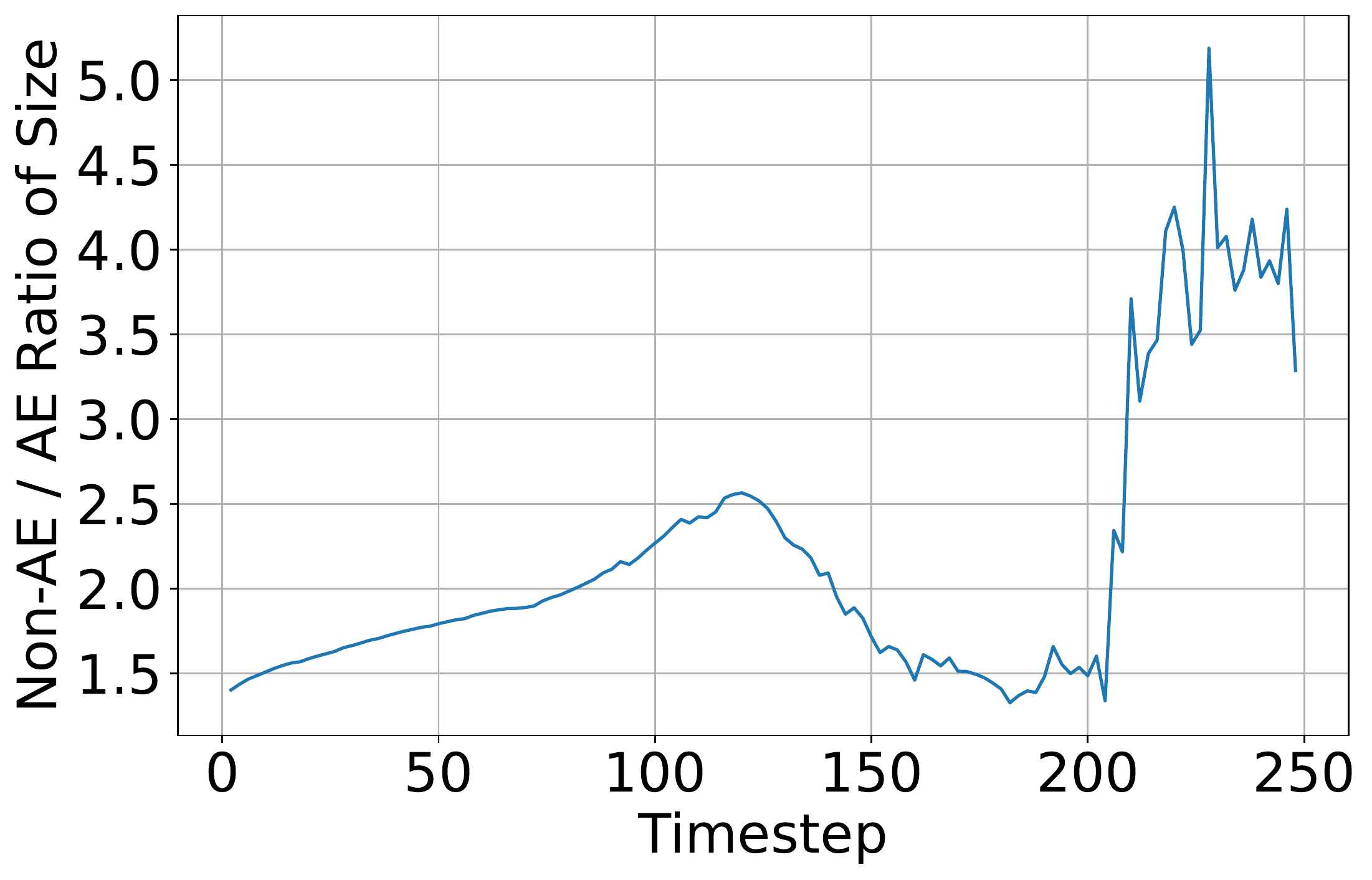}} \\
    \vspace{-0.8cm}
    \subfloat[PD errors]
    {\includegraphics[width=0.9\columnwidth]{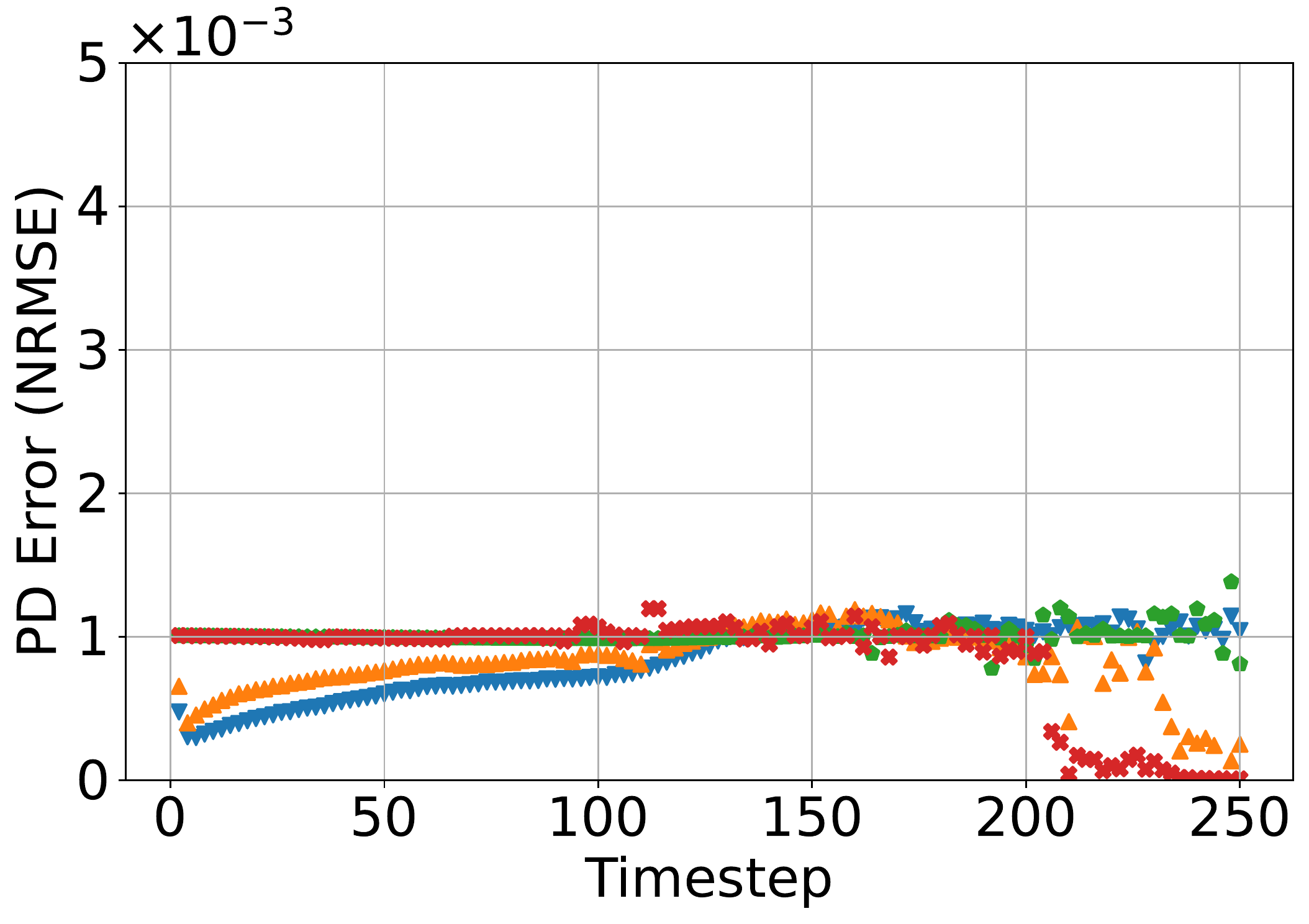}}
    \quad
    \subfloat[QoI errors]
    {\includegraphics[width=0.9\columnwidth]{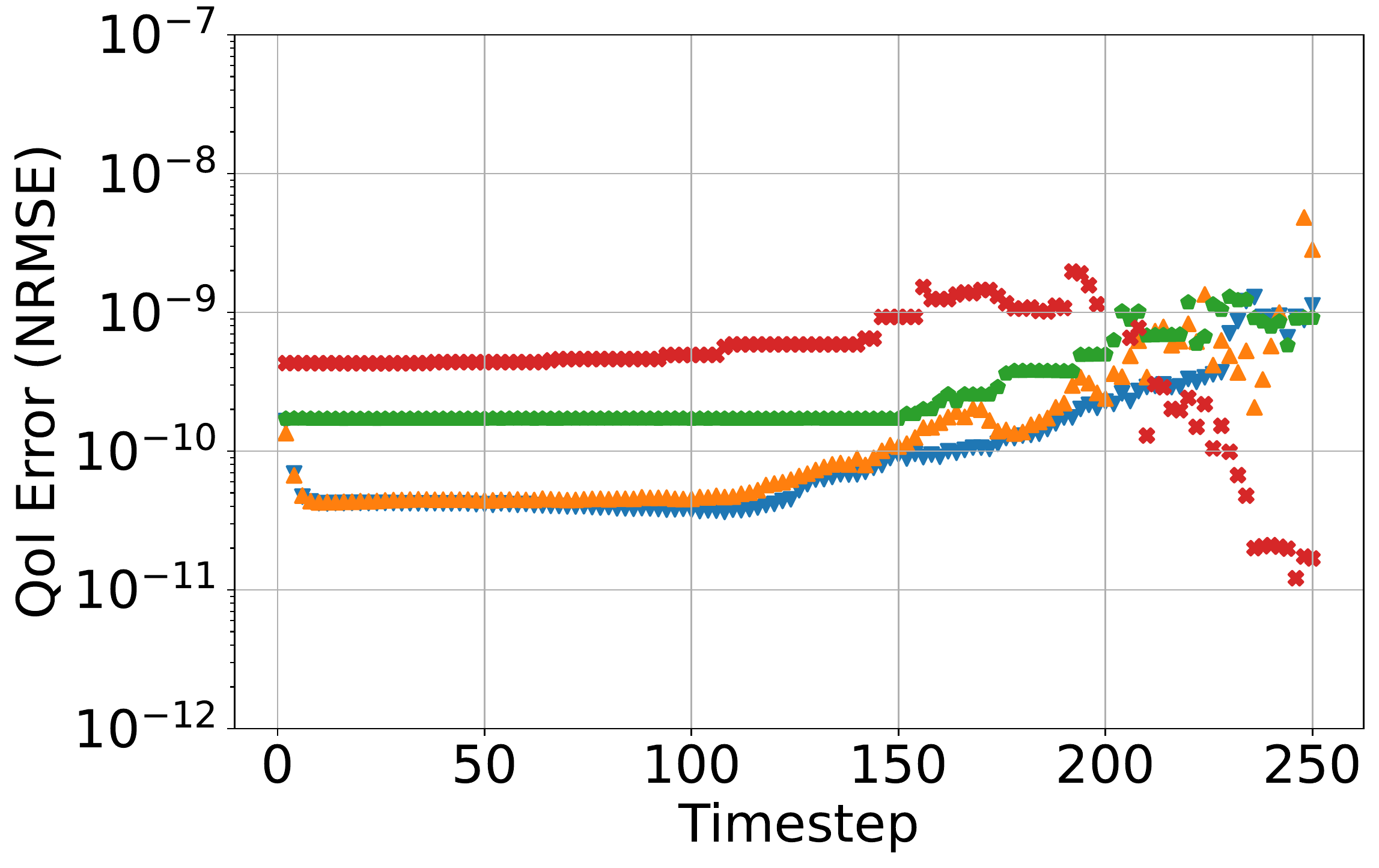}}
    \caption{(a) Compression ratio patterns with ITER simulation timestep, while keeping PD errors at $1e-3$, and QoI errors below $1e-8$. The $\lambda$s are represented using single precision floating point numbers. The data selection scheme used for training was COLRANDIND Paradigm 1 was used for choosing timesteps for full and incremental training. (b) Illustration showing PD errors are within the 1e-3 bound. (c) Illustration for QoI errors for the timesteps.}
    \label{fig:iterset}
    \vspace{-0.5cm}
\end{figure*}

\begin{figure}
    \centering
    \includegraphics[width=\columnwidth]{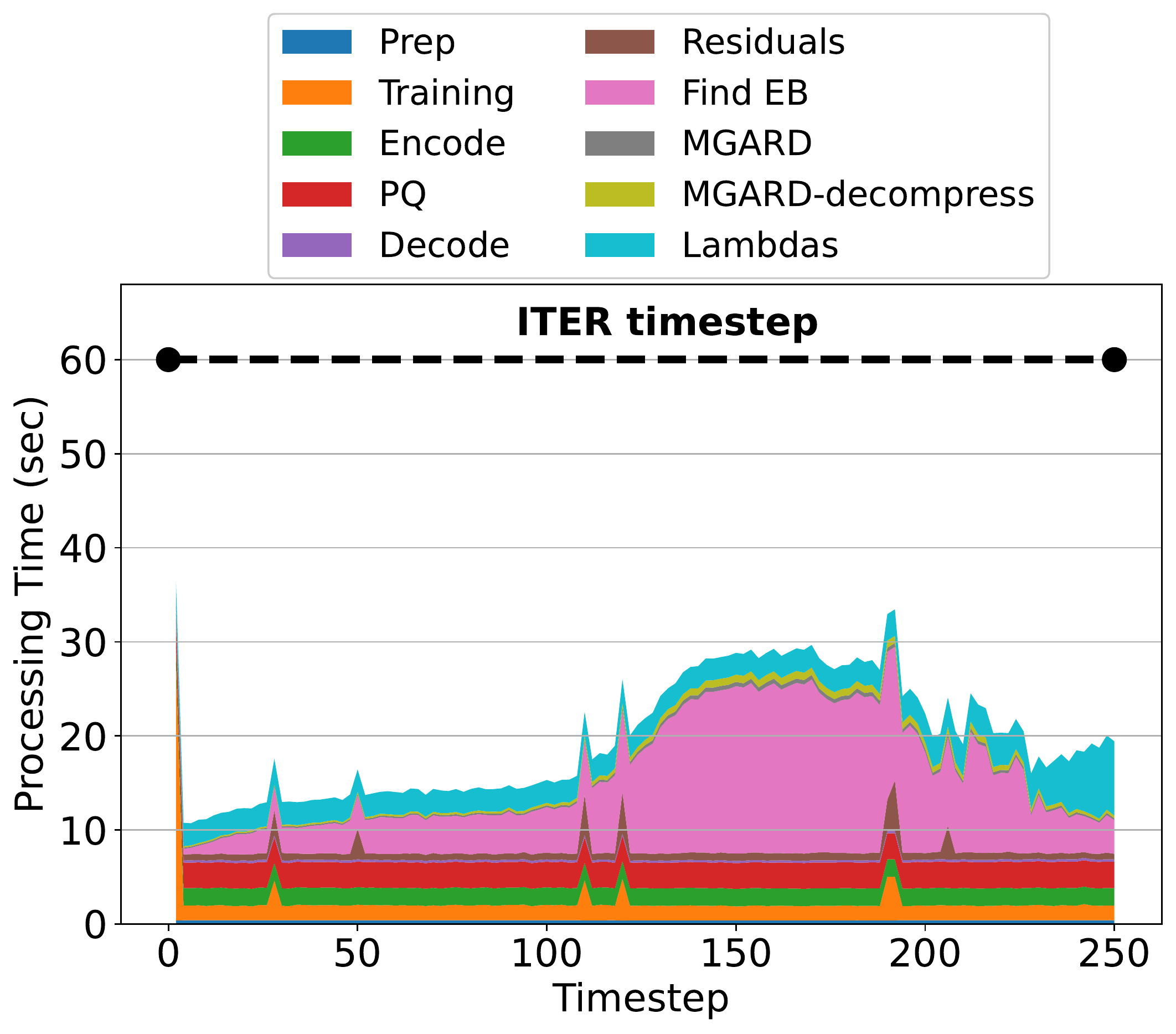}
    \caption{Breakdown of average time taken for compressing data for the ITER timesteps for ions and electrons using the AE pipeline. The orange spikes happen because the model is trained from scratch in those timesteps.}
    \label{fig:itertime}
\end{figure}
\eatme{
\begin{table}
    \centering
    \caption{Compression ratios for the three AE training paradigms are comparable for electrons.}
    \includegraphics[width=\columnwidth,trim={5cm 0cm 5cm 0cm},clip]{figures/paradigm_for_electrons.pdf}
    \label{tab:paradigme}
\end{table}
}
The QoI error distribution with post-processing allows us to use PD error as a yardstick to cap the compression ratio. Thus, when the PD error is $0.001$, the best compression pipeline for both ions and electrons is the AE-based pipeline presented in this paper with residual compression using MGARD and with post-processing using single precision $\lambda$s. This pipeline setting has a compression ratio of about $180$ for ions and $157$ for electrons. Compared to not using AE and only using MGARD with post-processing~\cite{hipc2022}, the compression ratio is higher by a factor of $1.67$ and $1.13$ for ions and electrons, respectively, for timestep $170$. If we consider the joint compression ratio (ions + electrons), then the data reduction by the AE pipeline for this timestep is $1.38$ times that achieved by the MGARD based non-AE pipeline. We compare the relative performance for all timesteps in Section~\ref{sec:iterexpt}. For the rest of the experiments, we stored $\lambda$s in single precision floating point format for better compression ratios for both the AE and the MGARD pipelines, and the setting (6), MG + $\lambda$ (float), was used to represent the non-AE pipeline.

Figure~\ref{fig:ec} also shows that the automatic determination of error bounds for MGARD residual processing is comparable to the user-specified bounds for ions and electrons. The automated error-bound determination makes our compression pipeline user-friendly. Because the user does not have to run MGARD multiple times to find the appropriate MGARD error bound for maximizing residual compression or settle for a conservative error bound that gives moderate reduction. Indeed, the user would have to settle for a conservative bound because, in the production environment, compression would run in lockstep with simulation, and there is no scope for any user intervention.

\subsection{Results on entire simulation} \label{sec:iterexpt}
This section discusses the AE paradigms used in training for multiple ITER timesteps, the compression ratio and error trends with timesteps, and the timing and resource requirements. The version of ITER simulation we used in this paper has $125$ timesteps, starting from $2$ up to $250$, in increments of $2$.

\subsubsection{AE Training Paradigms}
Table~\ref{tab:par} shows the three AE training paradigms we experimented with for compressing ITER data across different timesteps. Table~\ref{tab:paradigmi} shows the comparative compression ratios using the three paradigms for ions, and the trend is similar for electrons. All three paradigms are comparable, though as time progresses, it is better to update the AE model because the static model (Paradigm 2) starts to get a little worse. For our remaining experiments, we use Paradigm 3.
We use the COLRANDIND data selection scheme for training the AE. The AE pipeline uses the `AEM + $\lambda$ (float) + ab' setup, while the non-AE pipeline is `MG + $\lambda$ (float)', numbers 3 and 6 as described in Section~\ref{sec:errors}.

\subsubsection{Compression Ratio and Errors}
Figure~\ref{fig:iterset}(a) shows the individual compression ratios for each ITER step for ions and electrons while using the AE pipeline and the non-AE pipeline. Figure~\ref{fig:iterset}(b) plots the ratio of the compressed byte sizes (ions + electrons data) every timestep for AE versus non-AE pipelines. We found that the ratio is between 1.33 and 5.19. Summing up the sizes of compressed data for ions and electrons for all timesteps, we get a compression ratio of $332$ using the AE pipeline and $153$ using the MGARD pipeline. For each of the methods presented, we  keep the PD error close to $0.001$.

\eatme{
We try to keep the PD error close to $0.001$ to make the comparison of compression ratios fair. Keeping the error bound as close as possible to $0.001$ is automated in the AE pipeline ('Find EB' stage). For the non-AE pipeline, we use a wrapper written in bash script to run the compression multiple times using binary search on PD error to find an error bound for which the PD error is as close as possible to $0.001$. Thus, the rounded PD error is close to $0.001$ for all cases, as shown in Figure~\ref{fig:iterset}(c), while the QoI errors using single-precision floating-point are all below $10^{-8}$ as shown in Figure~\ref{fig:iterset}(d). 
}

\subsubsection{Performance}
Figure~\ref{fig:itertime} presents the time to compress ion and electron data using the AE pipeline for all ITER simulation timesteps. The time-consuming pipeline stages are: automatically finding MGARD error bound for the residuals, training the AE model from scratch, post-processing for $\lambda$~computation, PQ, and encoding the images using the AE model. Training time spikes when we train from scratch (timestep 2). Finding error bounds takes more time in the turbulent regime. The remaining steps take little time. These remaining steps are prep (for setting up training data and data loaders), decoding the latent vectors, preparing the residuals, and running MGARD compression and decompression routines. The total training time for ITER ions and electrons is less than 30 seconds, which is well within the time bound of 1 minute for simulating an ITER timestep. Therefore, we can easily integrate the compression step in situ with ITER simulation, so that data generated from the previous timestep is compressed in parallel with the simulation of the current step.

\subsubsection{Resources}
The resources required to perform the compression were 32 computational units (each unit corresponding to 1 Summit node consisting of 6 GPUs and 42 cores) for achieving a total time of less than 30 seconds. The time required for conducting the simulation on 1024 computational units and then writing the data is 1 minute combining both ions and electrons. Thus, the total resource overhead for the compression is about 3.125\% of the resources required for simulation. The compression step can be pipelined with the XGC simulation (of course, by using additional resources) without impacting the overall time requirements for simulation, assuming that the time to move the data from simulation computational units to compression computational units is relatively small (which is indeed the case). 

\section{Related Work\label{sec:related}}
Data compression for scientific applications is an active field of research. There are primarily two types of compression techniques, lossless and lossy. The compression technique is lossless if original data can be recovered upon decompression. While lossless compression~\cite{lossless1, zstd, osti}  preserves the original data, the compression achieved is significantly smaller than that of a lossy compression technique. Hence lossy techniques that can guarantee bounds on the PD errors are the most popular compression techniques used by scientists. Examples of lossy compression techniques are SZ~\cite{SZ_1, tao2017significantly, SZ_3}), which is based on prediction, ZFP~\cite{lindstrom2014fixed}, which is based on block transform, and MGARD~\cite{MGARD_1, MGARD_2, MGARD_3}, which employs a multilevel decomposition scheme. The main focus of these error-bounded lossy compression algorithms is point-wise errors on the primary data, and they provide a guaranteed error bound that the user can specify. However, these lossy compression schemes do not provide any bounds on the QoIs, which are critical for accurate data analysis. For example, erroneous first- and second-order moments in XGC uncompressed datasets could lead to a spurious fusion analysis. MGARD is currently the leading error-controlled lossy compression technique with bounded errors on the linear QoI. Other state-of-the-art lossy compression techniques such as SZ~\cite{SZ_1, tao2017significantly, SZ_3}  and ZFP~\cite{lindstrom2014fixed} typically suffer from low fidelity on the reconstructed data at high compression ratios (100+). 
Because the error bound needs to be set large to achieve a high compression ratio, this leads to considerable errors relative to the PD. Moreover, these techniques cannot reduce the errors in the QoIs directly. In the preliminary version of our paper~\cite{hipc2022}, we used a lossy compressor for data compression. Then we solved a constraint satisfaction problem using the Lagrange method to improve the errors on linear and complex nonlinear QoIs. In particular, we developed an MGARD-based compression pipeline. We studied the differences in results in terms of compression ratios, errors, and execution time between the scenarios when only MGARD is used versus when our pipeline is used. This paper is an extension of~\cite{hipc2022}. We introduce an autoencoder to the compression pipeline and run extensive evaluations to set up the pipeline parameters. Further, the implementation in this paper is GPU-based (\cite{hipc2022} was CPU-based), which improves compression times. This paper treats both ion and electron data whereas~\cite{hipc2022} treated ion data only.

\eatme{
Lee et al.~\cite{jaemoon2} presented an error-bounded lossy compressor based on autoencoders for guaranteed PD errors with QoI preservation. This work is different from~\cite{jaemoon2} in the following ways. From a pipeline implementation point of view, our pipeline in this paper is more straightforward as it does not use the Tucker decomposition or the inverse Tucker decomposition stages. Furthermore, we do not apply PQ to the Lagrange multipliers ($\lambda$s), which better preserves QoI accuracy at the cost of compression. 
}

\section{Conclusion\label{sec:concl}}
This paper presented a physics-informed compression technique implemented as an end-to-end, scalable, GPU-based pipeline for scientific data compression. Our hybrid compression technique combined an autoencoder, an error-bounded lossy compressor to provide guarantees on raw data error, and a constraint satisfaction post-processing step to preserve the QoIs within a minimal error (generally less than floating point error). We demonstrated on an XGC-based simulation of ITER~\cite{iter} that by using our AE-based data compression pipeline, we could achieve a compression level of $332$ while ensuring that the error on the primary data is less than $10^{-3}$ and that of the QoIs is less than $10^{-8}$. The data size was 20.25~TB for the 125-timestep simulation ($\sim162$~GB per timestep for ions and electrons), and the compression time was $2491$ seconds ($\sim20$ seconds per timestep). In comparison, using a non-AE and MGARD-only compression pipeline~\cite{hipc2022}, the compression ratio was $153$ for the same levels of PD and QoI errors.

We needed a standard lossy compressor with error guarantees to process the residuals on some images for which the AE returned higher errors than the user-specified threshold. We used MGARD for compressing the residuals.
Although other primary data compressors, such as SZ and ZFP, may also be used in our data compression pipeline, we found MGARD superior because it has lower PD and QoI errors than SZ and ZFP for the same level of compression.

The additional resources required to perform the compression were a few percent of the simulation resources. Further, our compression approach is inherently parallel, and more resources can be easily used, reducing the total time for compression, should it be necessary. The compression step can be pipelined with the simulation (of course, by using the resources required for compression) without impacting the overall time requirements for simulation. This may be achieved by compressing the data generated from the previous timestep in parallel with the simulation of the current step.

All of the above characteristics of our approach make it highly practical for on-the-fly compression while guaranteeing error bounds on PD and on QoIs---critically important for scientists to have confidence when applying data reduction.
\bibliographystyle{abbrv}
\bibliography{main}
\begin{IEEEbiography} [{\includegraphics[width=1in,height=1.25in,clip,keepaspectratio]{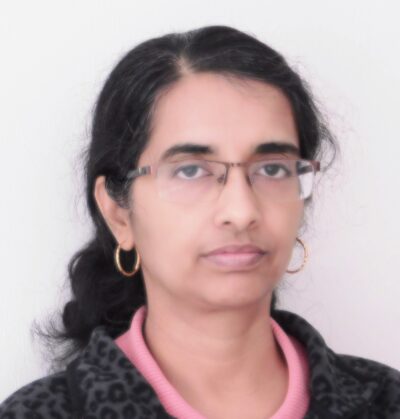}}] {Tania Banerjee}
Tania Banerjee is a research assistant scientist in Computer and Information Science and Engineering at the University of Florida. She obtained his Ph.D. in Computer Science from the University of Florida in 2012. She has done her M.Sc. in Mathematics from the Indian Institute of Technology, Kharagpur. Her research interests are data compression, high performance computing, video analytics, and intelligent transportation.
\end{IEEEbiography}
\begin{IEEEbiography} [{\includegraphics[width=1in,height=1.25in,clip,keepaspectratio]{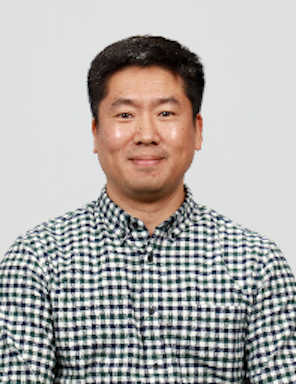}}] {Jong Choi}
Jong Youl Choi is a researcher working in the Discrete Algorithms Group, Computer Science and Mathematics Division, Oak Ridge National Laboratory (ORNL), Oak Ridge, Tennessee, USA. He earned his Ph.D. degree in Computer Science at Indiana University Bloomington in 2012 and his MS degree in Computer Science from New York University in 2004. His areas of research interest span data mining and machine learning algorithms, high-performance data-intensive computing, and parallel and distributed systems. 
\end{IEEEbiography}
\vspace{-1cm}
\begin{IEEEbiography} [{\includegraphics[width=1in,height=1.25in,clip,keepaspectratio]{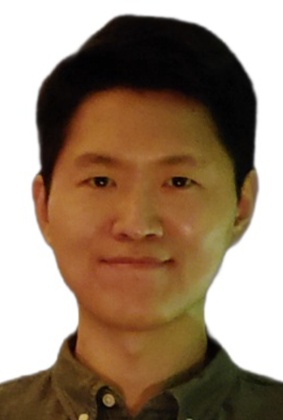}}] {Jaemoon Lee}
Jaemoon Lee is currently a Ph.D. student in the Department of Computer \& Information Science \& Engineering, University of Florida, Gainesville, FL. He has done his M.S. degree in CS from the University of Florida (2020). Prior to that, he worked as an Engineer for Mobile Communications Division in Samsung Electronics. His research interests are machine learning, physics-informed neural networks, and data compression.
\end{IEEEbiography}
\vspace{-1.3cm}
\begin{IEEEbiography} [{\includegraphics[width=1in,height=1.25in,clip,keepaspectratio]{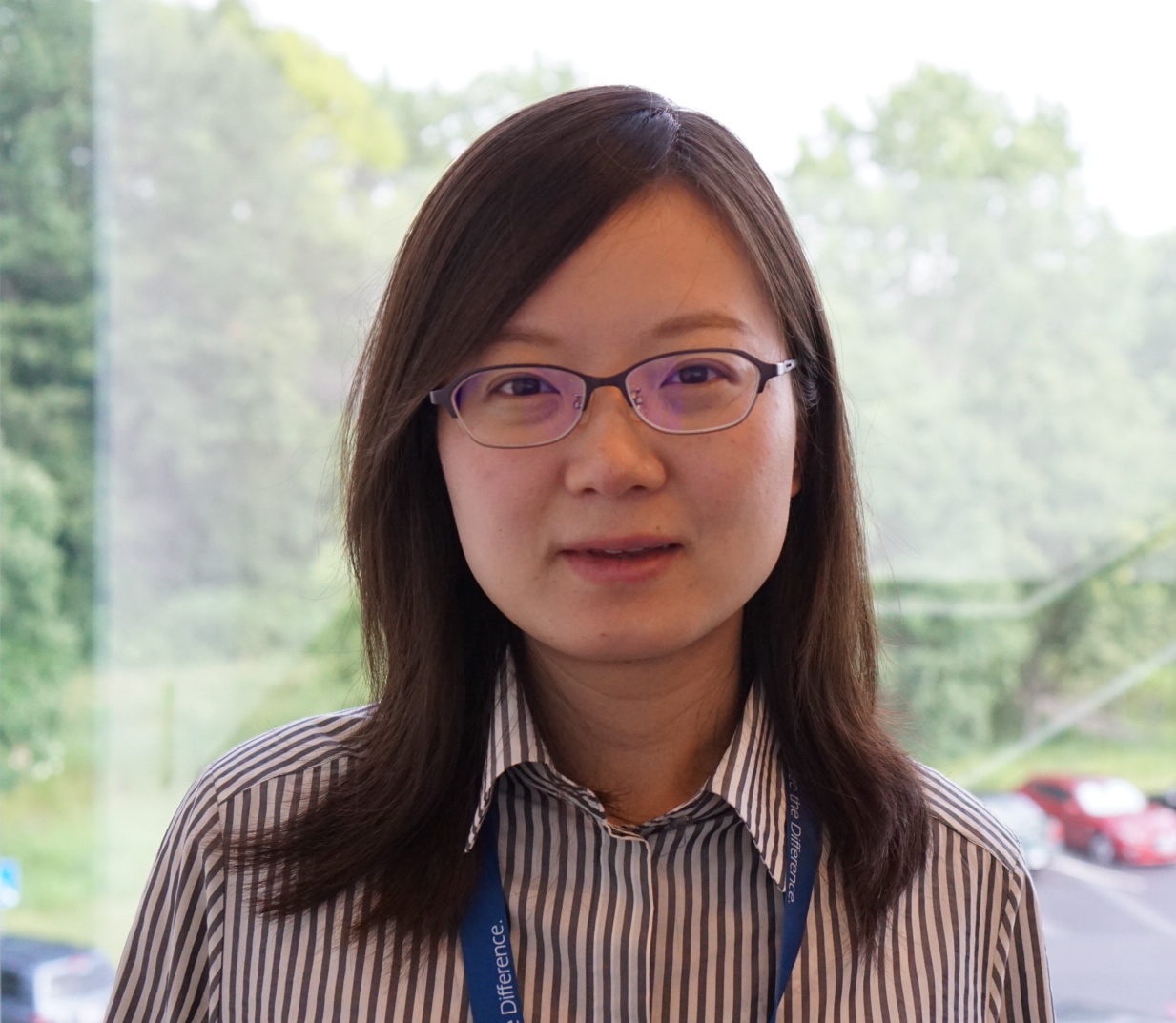}}] {Qian Gong}
Qian Gong is a Computer Scientist in the Computer Science and Mathematics Division at Oak Ridge National Laboratory. Prior to that, she worked as a Computer Science Researcher in the Network Research Group at Fermi National Accelerator Laboratory. She received her Ph.D. degree from Duke University (2017), a M.S. degree from University of Arizona (2013), and B.S. degree from Hefei University of Technology (2010). Her research interests are high-performance computing and scientific data management, with special emphasis in scientific data compression, parallel computing, and parallel I/O.
\end{IEEEbiography}
\vspace{-1cm}
\begin{IEEEbiography} [{\includegraphics[width=1in,height=1.25in,clip,keepaspectratio]{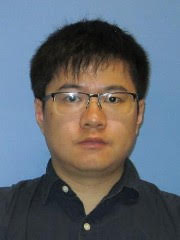}}] {Jieyang Chen}
Jieyang Chen assistant professor at University of Alabama at Birmingham (UAB). Before joining UAB, he was a computer scientist at Computer Science and Mathematics Division at Oak Ridge National Laboratory. He received his master and Ph.D. degrees in Computer Science from University of California, Riverside in 2014 and 2019. His research interests include high-performance computing, parallel and distributed systems, and big data analytics.
\end{IEEEbiography}
\vspace{-1cm}
\begin{IEEEbiography} [{\includegraphics[width=1in,height=1.25in,clip,keepaspectratio]{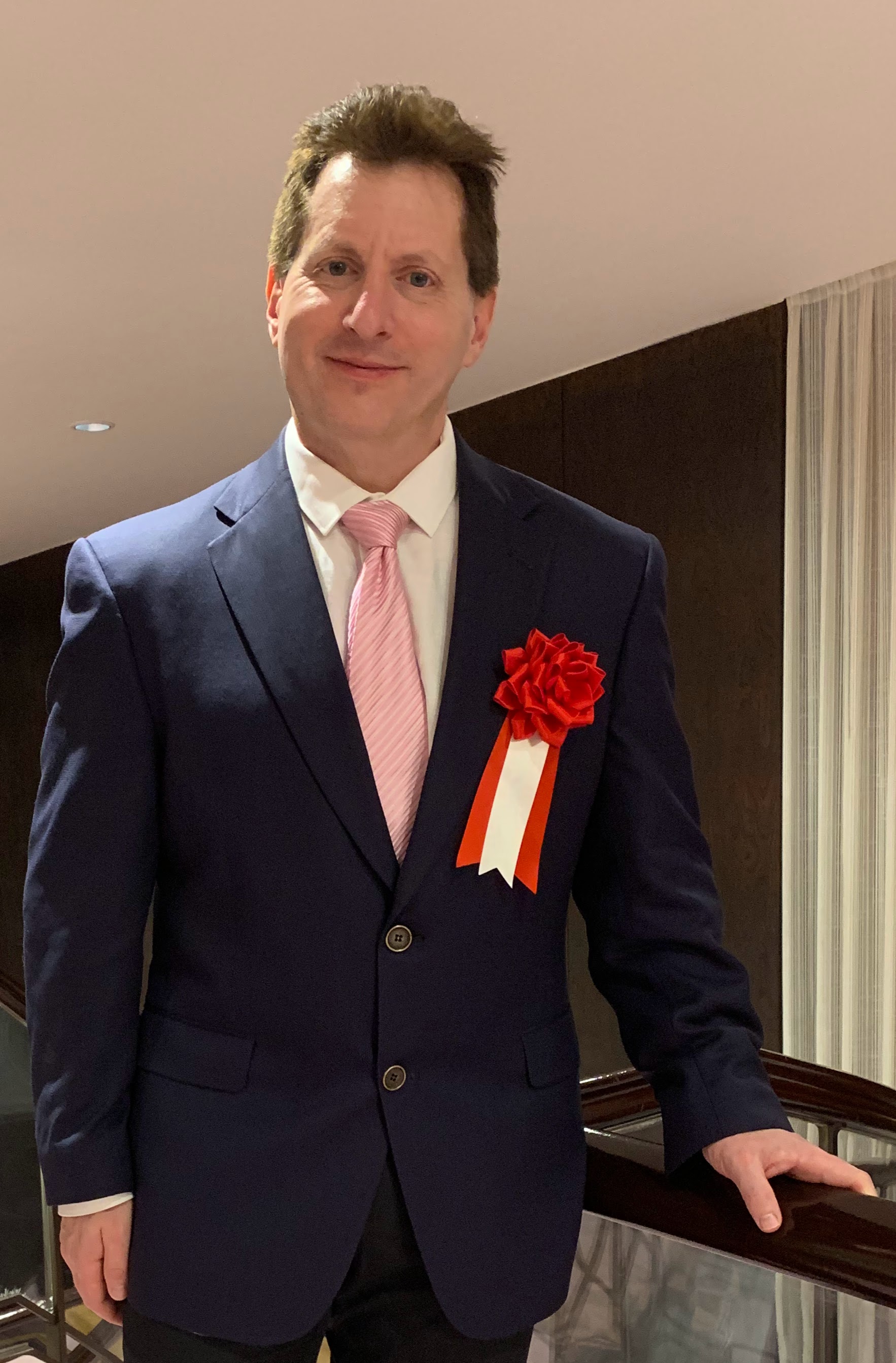}}] {Scott Klasky}
Scott A. Klasky is a distinguished scientist and the group leader for Workflow Systems in the Computer Science and Mathematics Division at the Oak Ridge National Laboratory. He also holds an appointment at the at  Georgia Tech University. He obtained his Ph.D. in Physics from the University of Texas at Austin (1994). Dr. Klasky is a world expert in scientific computing, scientific data reduction, and scientific data management, co-authoring over 300 papers.
\end{IEEEbiography}
\vspace{-1.2cm}
\begin{IEEEbiography}[{\includegraphics[width=1in,height=1.25in,clip,keepaspectratio]{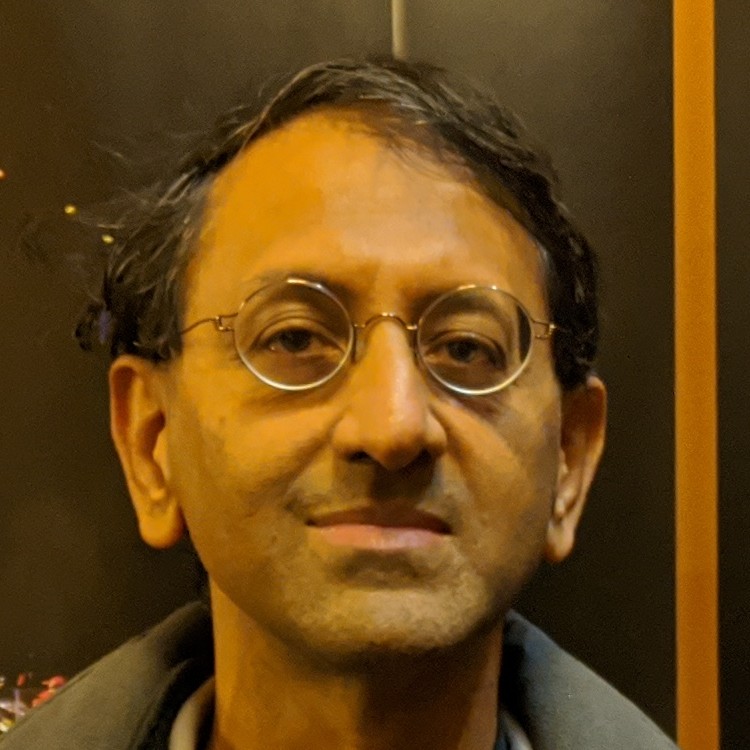}}] {Anand Rangarajan}
Anand Rangarajan is Professor, Dept. of CISE, University of Florida. His research interests are machine learning, computer vision, medical and hyperspectral imaging and the science of consciousness.
\end{IEEEbiography}
\vspace{-1.3cm}
\begin{IEEEbiography}[{\includegraphics[width=1in,height=1.25in,clip,keepaspectratio]{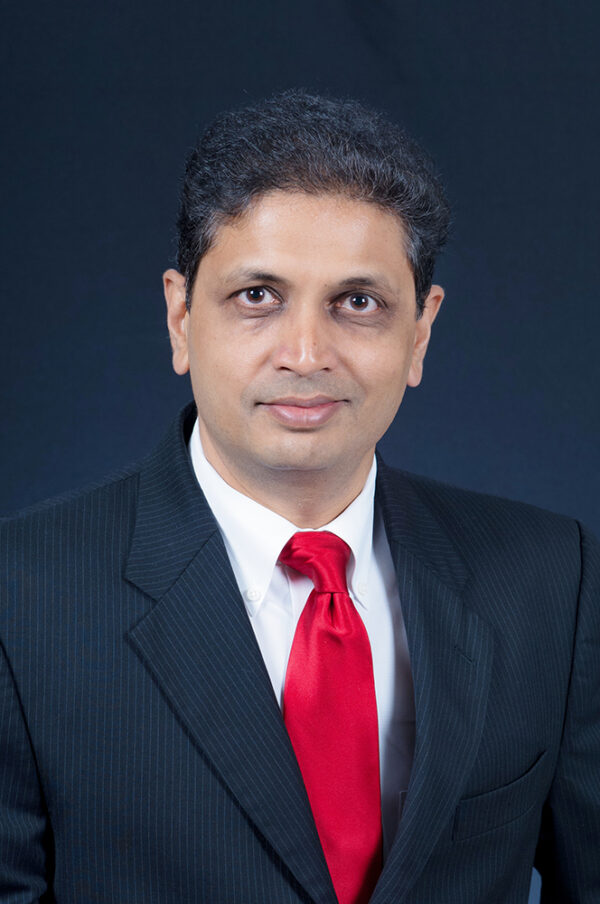}}] {Sanjay Ranka}
Sanjay Ranka is a Distinguished Professor in
the Department of Computer Information Science and
Engineering at University of Florida. His current research interests are high performance computing and big data science with a focus on applications in CFD, healthcare and transportation. He has co-authored four books, 290+ journal and refereed conference articles. He is a Fellow of the IEEE and AAAS. He is an Associate Editor-in-Chief of the Journal of Parallel and Distributed Computing and an Associate Editor for ACM Computing Surveys, Applied Sciences, Applied Intelligence, IEEE/ACM Transactions on Computational Biology and Bioinformatics.
\end{IEEEbiography}
\vspace{-1cm}
\end{document}